# A Lightweight and Robust Framework for Real-Time Colorectal Polyp Detection Using LOF-Based Preprocessing and YOLO-v11n

Saadat Behzadi[a], Danial Sharifrazi[b], Bita Mesbahzadeh[c], Javad Hassannataj Joloudari[d,e], Roohallah Alizadehsani[b*]

[a] Department of Electronic Engineering, University of Bologna, Bologna, Italy
[b] Institute for Intelligent Systems Research and Innovations (IISRI), Deakin University, Geelong, Australia
[c]Department of Information Technology Engineering, Faculty of Industrial and Systems Engineering, Tarbiat Modares University, Tehran, Iran
[d]Department of Computer Engineering, Faculty of Engineering, University of Birjand, Birjand, Iran
[e]Department of Computer Engineering, Technical and Vocational University (TVU), Tehran, Iran

**Abstract**.

Timely and accurate detection of colorectal polyps played a crucial role in diagnosing and preventing colorectal cancer, a major cause of mortality worldwide. This study introduced a new, lightweight, and efficient framework for polyp detection that combined the Local Outlier Factor (LOF) algorithm for filtering noisy data with the YOLO-v11n deep learning model. An experimental study leveraging deep learning and outlier removal techniques across multiple public datasets.The proposed approach was tested on five diverse and publicly available datasets: CVC-ColonDB, CVC-ClinicDB, Kvasir-SEG, ETIS, and EndoScene. Since these datasets originally lacked bounding box annotations, we converted their segmentation masks into suitable detection labels. To enhance the robustness and generalizability of our model, we apply 5-fold cross-validation and remove anomalous samples using the LOF method configured with 30 neighbors and a contamination ratio of 5%. Cleaned data are then fed into YOLO-v11n, a fast and resource-efficient object detection architecture optimized for real-time applications. We trained the model using a combination of modern augmentation strategies to improve detection accuracy under diverse conditions. Our approach significantly improved polyp localization performance, achieving a precision of 95.83%, recall of 91.85%, an F1-score of 93.48%, an mAP@0.5 of 96.48%, and an mAP@0.5:0.95 of 77.75%. Compared to previous YOLO-based methods, our model demonstrated enhanced accuracy and efficiency.
These results suggested that the proposed method is well-suited for real-time colonoscopy support in clinical settings. Overall, the studied underscores how crucial data preprocessing and model efficiency are when designing effective AI systems for medical imaging.

## 1. Introduction

Colorectal cancer (CRC) remained one of the most common and deadly forms of cancer globally. According to recent statistics, approximately 1.9 million new cases and 900000 deaths were recorded worldwide in 2020, ranking third in incidence and second in cancer-related mortality. [1, 2]. The development of CRC often began with benign adenomatous polyps, which can gradually transform into malignant tumors if not detected and removed promptly. [3-5]. This adenoma-to-carcinoma sequence underscores the importance of early detection and intervention.

Colonoscopy is currently considered the gold standard for detecting and removing polyps. However, studies have shown that up to 26% of polyps are missed during routine colonoscopies, especially small, flat, and mucosa-camouflaged lesions. [6-8]. These shortcomings may result from endoscopies fatigue and suboptimal image quality. Therefore, the integration of computer-aided detection (CADe) systems into colonoscopy workflows has gained significant attention for their potential to improve diagnostic accuracy and reduce human error. [9].

In response to this need, artificial intelligence particularly deep learning has demonstrated promising capabilities in medical imaging, where convolutional neural networks (CNNs) have been widely adopted for feature extraction and classification; building on this foundation, YOLO (You Only Look Once) has emerged as a breakthrough in object detection, offering real-time performance and high accuracy that make it especially well-suited for colonoscopy applications [10-13].
The YOLO series has undergone continuous evolution, with newer versions such as YOLOv5, YOLOv8, and YOLOv11 offering improvements in architecture, speed, and feature extraction techniques. [14-17]. However, challenges like poor image quality and noise from endoscopic equipment still hindered detection performance. [18].
Beyond YOLO-based detection, prior research has explored a broad range of methods. Early techniques utilized hand-crafted features such as color, texture, and shape, coupled with conventional classifiers like SVMs and decision trees [20, 21]. Although effective in controlled settings, these approaches lacked generalization. The shift to deep learning, particularly CNN-based architectures like FCNs and U-Net, greatly improved segmentation performance [22-25]. Advanced networks such as PraNet, AMNet, and FANet further refined predictions through modules like reverse attention, parallel-partial decoders, and feature enhancement mechanisms. [24, 26, 27].
In addition, ensemble and hybrid networks such as DDANet and DeiT-ResNet demonstrated superior performance through multi-scale feature aggregation. [28-31]. YOLOv3 to YOLOv5 models also incorporated innovations such as feature fusion and self-attention modules, achieving notable gains in small polyp detection. [32-35]. YOLOv8 introduced new modules like Reverse Attention with Semantic context (RA-S), enhancing both local and global feature learning. [13, 25].
GAN-integrated data synthesis models demonstrated enhanced generalization and reduced overfitting when trained on a combination of real and synthetic data. [36].
Despite their advancements, these models still encountered challenges, including difficulty detecting subtle and small polyps and dependency on high-quality training data. [37-40]. Additionally, many approaches lacked proper handling of noisy samples, which can distort model training and lead to reduced performance.
To mitigate these issues, researchers have proposed enhancements such as attention mechanisms integrated into YOLO architectures, enabling the model to better focus on relevant image regions while suppressing irrelevant background noise. [4, 37, 38, 41, 42]. These techniques improved detection accuracy, particularly for ambiguous and hard-to-detect polyps, while maintaining real-time inference speeds suitable for clinical settings.
Nonetheless, the role of data quality in model training has often been overlooked. Public colonoscopy datasets were typically small, imbalanced, and included images with occlusions, noise, and poor visibility. Such anomalies can negatively impact model performance.
To resolve persistent challenges such as false positives, poor generalization across institutions, and sensitivity to imaging conditions, we proposed the LOF-YOLOv11n framework. This novel approach combined outlier removal using the LOF algorithm [39] with the YOLOv11n object detection model [40], integrating robust data cleaning, efficient architecture, and rigorous evaluation. As a result, it represented a practical solution capable of achieving high detection performance under real-world clinical constraints.
LOF was a density-based anomaly detection method that identified data points with abnormal local density, making it effective in filtering noisy training images. By removing outlier samples before training, the model could learn from higher-quality data, which led to improved robustness and accuracy. Following the LOF preprocessing, the clean dataset is used to train YOLOv11n. A lightweight and optimized model designed for real-time detection tasks with constrained computational resources [40]. YOLOv11n included key architectural enhancements such as the C3k2 block and C2PSA attention module, which improve accuracy while maintaining low inference latency.
The model is trained and evaluated on five publicly available colonoscopy datasets: CVC-ColonDB, CVC-ClinicDB, Kvasir-SEG, ETIS, and EndoScene. [39, 43]. As these datasets are originally segmentation-based, we converted their pixel-level masks into bounding boxes for object detection. A five-fold cross-validation strategy is used to ensure balanced and reliable evaluation across diverse datasets and anatomical variations.
Our evaluation metrics included precision, recall, F1-score, and mean Average Precision (mAP) at IoU thresholds of 0.5 and 0.5:0.95. The proposed LOF-YOLOv11n framework outperformed both the baseline YOLOv11n and previous YOLO-based models, such as YOLOv8m and YOLOv10, demonstrating its superiority in detecting challenging polyp cases. [44, 45]. Specifically, the model achieved a precision of 95.83%, a recall of 91.85%, an F1-score of 93.48%, and mAP@0.5 of 96.48%, validating its effectiveness and practical applicability in clinical environments.
The main contributions of our study are:

- Integration of LOF with YOLOv11n for enhanced training data quality.
- Use of segmentation-to-bounding box conversion for compatibility with object detection frameworks.
- Application of YOLO-v11n Nano for real-time performance with reduced memory and computational overhead.
- 5-fold cross-validation across five benchmark datasets to ensure reliability and generalizability.

- Significant improvements over previous models in terms of detection precision, recall, and computational efficiency.

* Corresponding author: Roohallah Alizadehsani

The Rest of the study is organized as Follows: Section 2 presented a review of related work in polyp detection and deep CADe systems. Section 3 provided the evaluation results and a detailed explanation of dataset preprocessing, the LOF algorithm, the YOLOv11n architecture, and the training pipeline. Section 4 focused on the presentation and analysis of experimental results and performance metrics. Section 5 concluded the study and outlined proposed directions for future research, and Section 6 checked future work.

## 2. Related work

Among the previous research efforts, computer vision and deep learning-based methods have been extensively applied for automatic colorectal polyp detection. Nevertheless, this task remained highly challenging due to the visual similarity of polyps to surrounding mucosal tissues, variations in shape and size, and the influence of imaging artifacts such as specular highlighted or motion blur. In this section, recent studies on colorectal polyp detection are examined from the perspectives of deep learning approaches.

Wan et al. [38] proposed a YOLOv5 model integrated with an attention mechanism using the Kvasir-SEG dataset. Their approach achieved a precision of 91.50%, a recall of 89.90%, and an F1-score of 90.70%. The method exploited attention layers to capture contextual features more effectively, demonstrating the impact of attention-based enhancements in real-time detection frameworks.

Qadir et al. [46] introduced the MDeNetplus (F-CNN) framework, evaluated on CVC-ColonDB. Their method achieved 88.35 % precision, 91 % recall, and 89.65 % F1-score. By using multi-scale convolutional features, the model was able to capture both local and contextual patterns, enhancing the detection of small and irregular polyps.

Shin et al. [47] applied a CNN-based approach combining patch classification with bounding-box localization, trained across multiple datasets including CVC-ClinicDB, ETIS-LARIB, and ASU-Mayo. Their model reached 92.71 % precision and 90.82 % recall, confirming the strength of CNN architectures in spatial localization tasks. The patch-based methodology allowed robust detection of polyps within complex clinical images.

Zhou et al. [48] proposed a method for polyp detection and radius measurement in the small intestine using video capsule endoscopy. Their approach utilized geometric and texture features to accurately identify polyps and measure their radii, aiding clinicians in faster and more precise diagnosis.

Sahoo et al. [49] explored real-time detection by implementing YOLOv11n on the Kvasir-SEG dataset. Their model achieved 91.76 % precision, 90.29 % recall, and 91.01 % F1-score. This lightweight adaptation demonstrated that YOLO-based frameworks can maintain high accuracy while being optimized for speed and deployment in resource-constrained environments.

Butler et al. [50] proposed a lightweight tracker combined with detection on the CVC-ClinicDB dataset. Their method achieved balanced performance, with 94.80 % across precision, recall, and F1-score. The inclusion of tracking modules reduced false negatives in sequential frames, making it especially suitable for endoscopic video analysis.

Jha et al. [51] conducted an extensive evaluation of ResUNet++, integrating Conditional Random Fields (CRF) and Test-Time Augmentation (TTA), across multiple benchmark datasets including CVC-ColonDB, CVC-ClinicDB, ETIS-LARIB, and Kvasir-SEG.Their model achieved strong segmentation performance, reporting mean Dice coefficient (F1-score) **of** 0.923, precision of 0.931, and recall of 0.914 on the Kvasir-SEG dataset, and Dice = 0.871, precision = 0.882, recall = 0.865 on the CVC-ClinicDB dataset. These results demonstrated that the combination of CRF post-processing and TTA improved boundary accuracy and overall generalization compared to the baseline ResUNet++ model.

Yeung et al. [52] proposed a lightweight U-Net with a focus mechanism, evaluated on Kvasir-SEG and CVC-ClinicDB. Using segmentation metrics, their method achieved mean Dice (DSC) = 0.941 on CVC-ClinicDB and DSC = 0.910 on Kvasir-SEG. Their results showed that architectural improvements (dual attention gating, hybrid focal-Tversky loss) effectively improve segmentation accuracy across diverse endoscopic image sources.

Lalinia et al. [25] introduced YOLOv8m enhanced with focus mechanisms across multiple datasets, including Kvasir-SEG, CVC-ClinicDB, CVC-ColonDB, ETIS, and EndoScene. Their method achieved 95.60 % precision, 91.70 % recall, and 92.40 % F1-score, showing significant improvements due to the incorporation of attention and focus modules. Another variant, YOLOv8n, achieved 95.20 % precision, 90.30 % recall, and 91.20 % F1-score, validating the scalability of lightweight YOLO versions for large-scale deployment.

Finally, our proposed method combined Local Outlier Factor (LOF)-based preprocessing with YOLOv11n detection. By leveraging LOF for robust outlier removal and YOLOv11n for lightweight, real-time detection, the framework was evaluated on Kvasir-SEG, CVC-ClinicDB, CVC-ColonDB, ETIS, and EndoScene datasets. The results demonstrated 95.83 % precision, 91.85 % recall, and 93.48 % F1-score, outperforming many prior studied in terms of balanced metrics. The integration of preprocessing with outlier removal contributed to more reliable detection under challenging imaging conditions, representing a practical and scalable solution for real-world clinical use.

## 3. Methods

This study presented a structured methodology for detecting polyps in colonoscopy images using a combination of outlier detection and a lightweight deep learning model. The method is designed to ensure efficiency, accuracy, and robustness while minimizing computational costs. The proposed pipeline followed a series of logical steps, each tailored to enhance data quality and detection precision. The pseudocode of the pre-trained model is presented below.

```
Algorithm: Polyp Detection in Colonoscopy Images using LOF-YOLO-v11n
Input:
   - Colonoscopy Datasets containing 2,248 images
        • Kvasir-SEG, CVC-ColonDB, CVC-ClinicDB, EndoScene, ETIS
Output:
   - Predicted bounding box annotation for all test images
   - Mean Evaluation metrics over 5-fold cross-validation
        • Precision, Recall, F1-score, mAP

1.   # Step 0: START

2.   # Step 1: Generate Labels and Prepare Dataset
3.   For each dataset IN datasets:
4.        For each Colonoscopy image in the IN dataset:
5.             Load the segmentation mask.
6.             Generate a bounding box annotation from the mask.
7.             Save the generated bounding box annotation.

8.   # Step 2: Split the Dataset and apply 5-fold cross-validation
9.   For fold in range(k=5):
10.       Split the dataset randomly into:
11.            65% Training
12.            15% Validation
13.            20% Testing

14.            # Step 3: Apply Outlier Detection (LOF) to the training, validation, and test sets
15.            For each set IN {Train, Validation, Test}:
16.                 For each Image IN the set:
17.                      Computing the LOF score
18.                 Sort LOF scores
19.                 Remove top samples corresponding to the contamination rate (outliers)
20.            # Step 4: Train YOLO-v11n
21.            Initialize YOLO-v11n with pre-trained weights
22.             Train and validate the model on cleaned training and validation sets.
23.            Save the best model weights.

24.            # Step 5: Predict Polyps Detection
25.            Load trained YOLO-v11n model.
26.            For each image IN the cleaned test image:
27.                 Predict bounding boxes for polyps.
```

28. Save predicted bounding box results.

29. # Step 6: Evaluate Proposed Method Performance
30. Compute evaluation metrics (Precision/ Recall/ F1-score/ mAP) on the Test set:
31. # Step 7: Compute the mean of evaluation metrics over 5 folds

32. # Step 8: END

Figure 1 illustrates the step-by-step process of the proposed methodology.

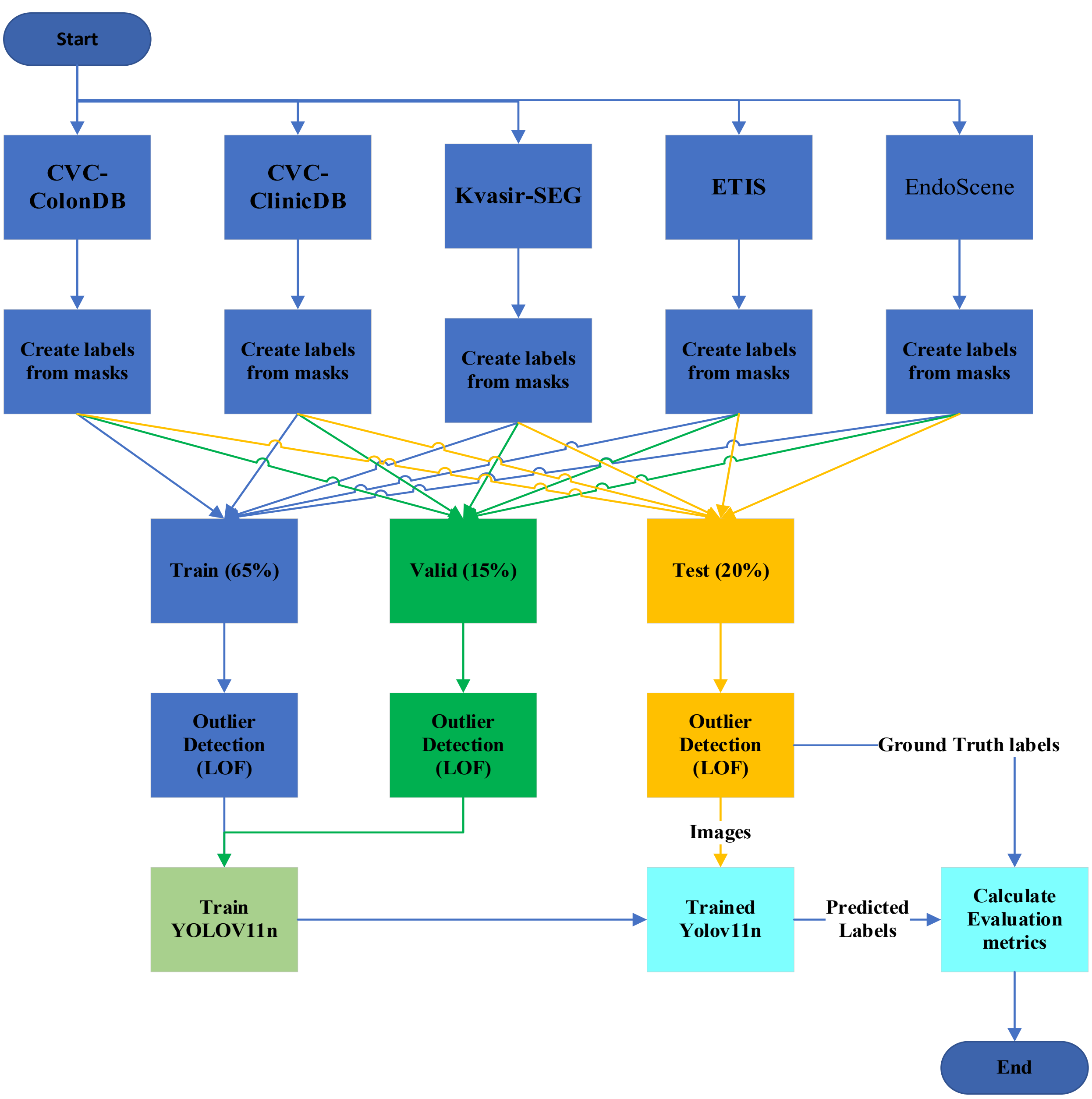

Figure 1. The proposed methodology.

Each stage is methodically designed to address specific challenges in polyp detection, from preprocessing and anomaly filtering to model training and performance evaluation.
The complete process includes the following key steps: - Step 0: Start - Step 1: Preprocess the Data and Create Labels from Masks - Step 2: Split Data into Train, Validation, and Test Sets (5-Fold Cross-Validation) - Step 3: Apply Outlier Detection (Local Outlier Factor) - Step 4: Train YOLO-v11n with Cleaned Training and Validation Datasets - Step 5: Predict Polyps' Detections Using the Trained Model - Step 6: Evaluate Model Performance - Step 7: End of the Process.
These steps are described in the subsections below.

### 3.1. Study design and Label Generation

Five publicly available polyp datasets are used in this study: CVC-ColonDB, CVC-ClinicDB, Kvasir-SEG, ETIS, and EndoScene. These datasets collectively contained 2,248 colonoscopy images in various sizes and formats. Since these datasets did not include standard bounding box annotations for polyps, segmentation masks are utilized to generate bounding box labels suitable for training YOLO-based models.
To ensure reliable evaluation and minimize bias, a 5-fold cross-validation approach is adopted. The datasets are divided as follows: 20% for testing, 15% for validation, and the remaining 65% for training. This split is performed randomly for each fold, and A summary of our research datasets is shown in Table 1.

**Table 1.** Summary of Original Datasets

| Dataset | Total number of images | Size | Type |
|---|---|---|---|
| CVC-ColonDB | 380 | 574 × 500 | PNG |
| CVC-ClinicDB | 612 | 384 × 288 | PNG |
| Kvasir-SEG | 1,000 | 487 × 332 - 1920x1072 | JPG |
| ETIS | 196 | 1225 × 966 | PNG |
| EndoScene | 60 | 574 × 500 | PNG |

During training of the polyp detection model, we used a diverse set of colonoscopy images annotated with bounding boxes around polyps. Sample batches included figure 2, figure 3, figure 4, figure 5, figure 6, and figure 7. These batches contained polyps with varying shapes, sizes, and appearances, captured under different imaging conditions.
The bounding boxes highlighted polyps in both clear and challenging scenarios, such as low contrast, partial occlusion, or the presence of fluids. For instance, figures 5 and 6 included complex and irregular polyps, while figure 7 features high-contrast images with detailed surface textures. This diversity ensured the model learns robust polyp localization across real-world clinical variations.

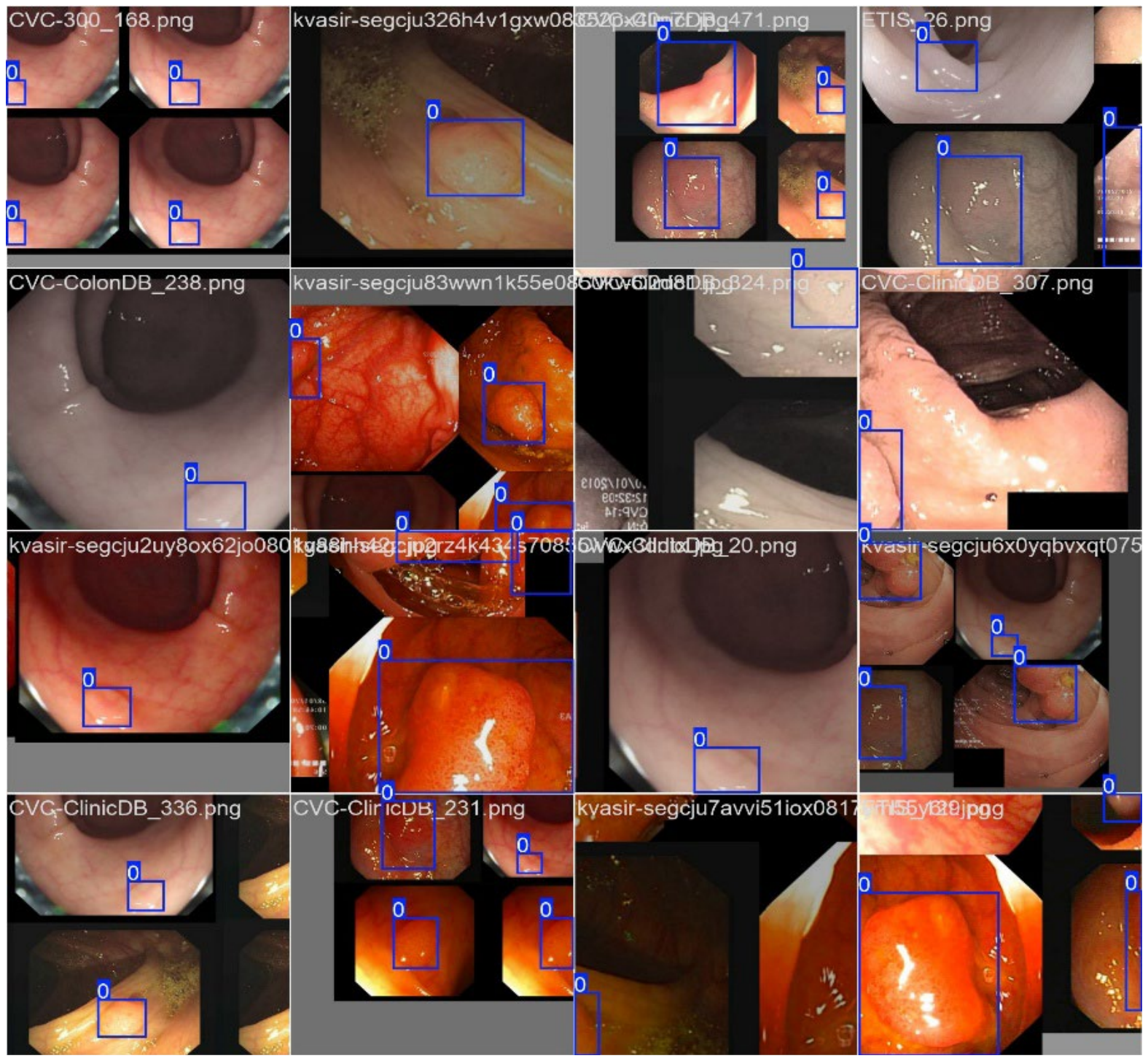

Figure 2. Sample images from Training Batch 0 for Polyp Detection.

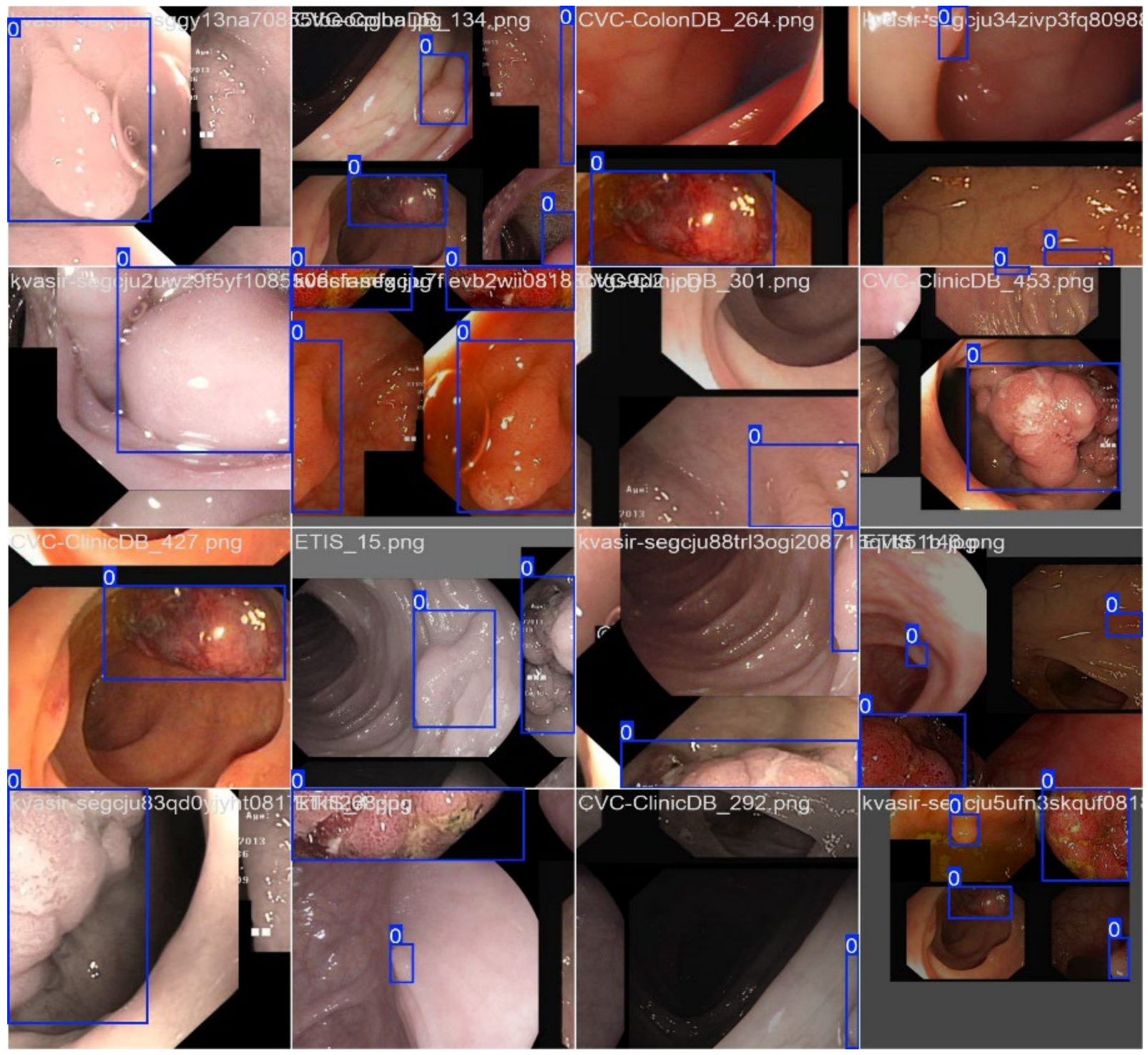

Figure 3. Sample images from Training Batch 1 for Polyp Detection.

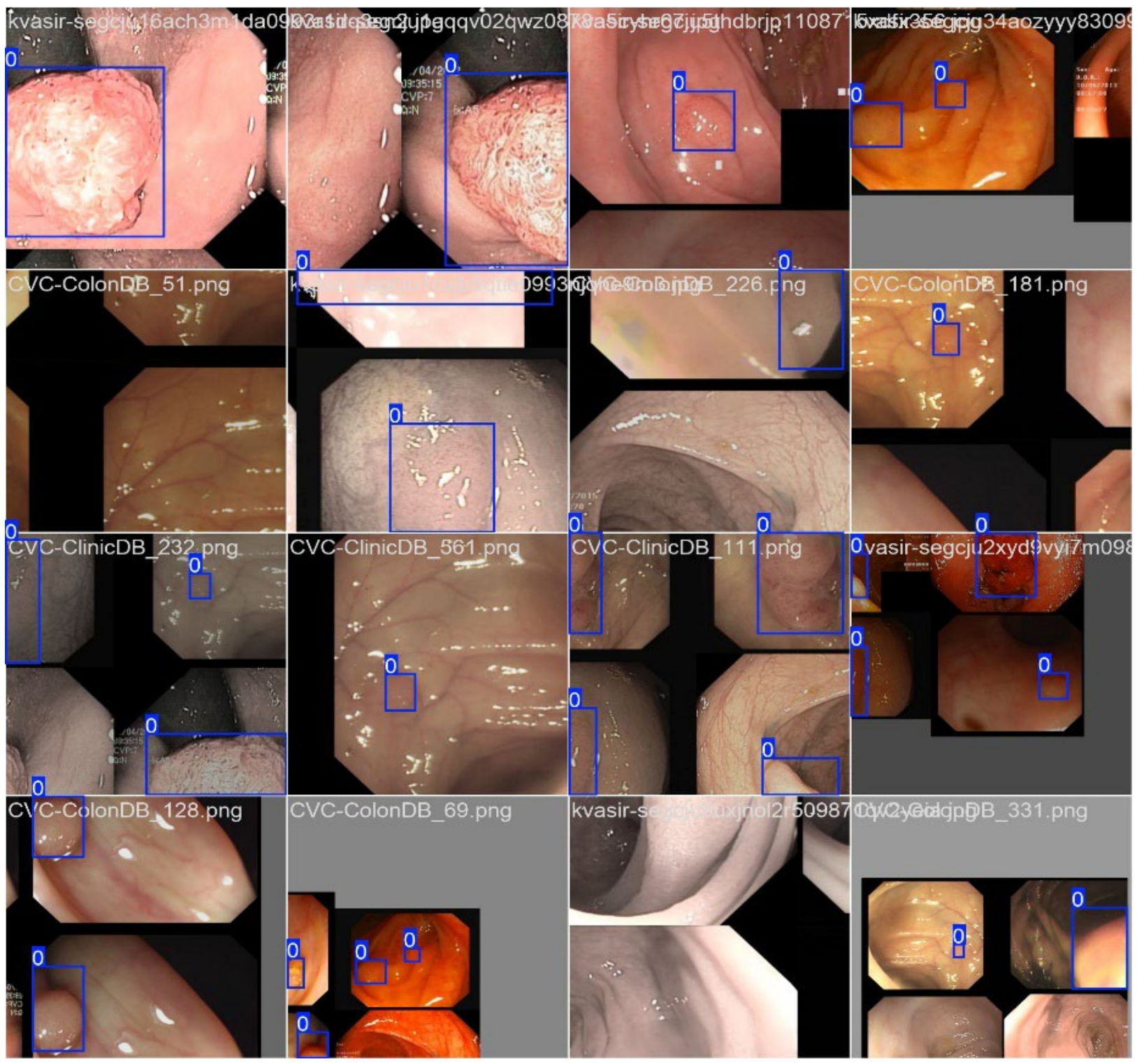

Figure 4. Sample images from Training Batch 2 for Polyp Detection.

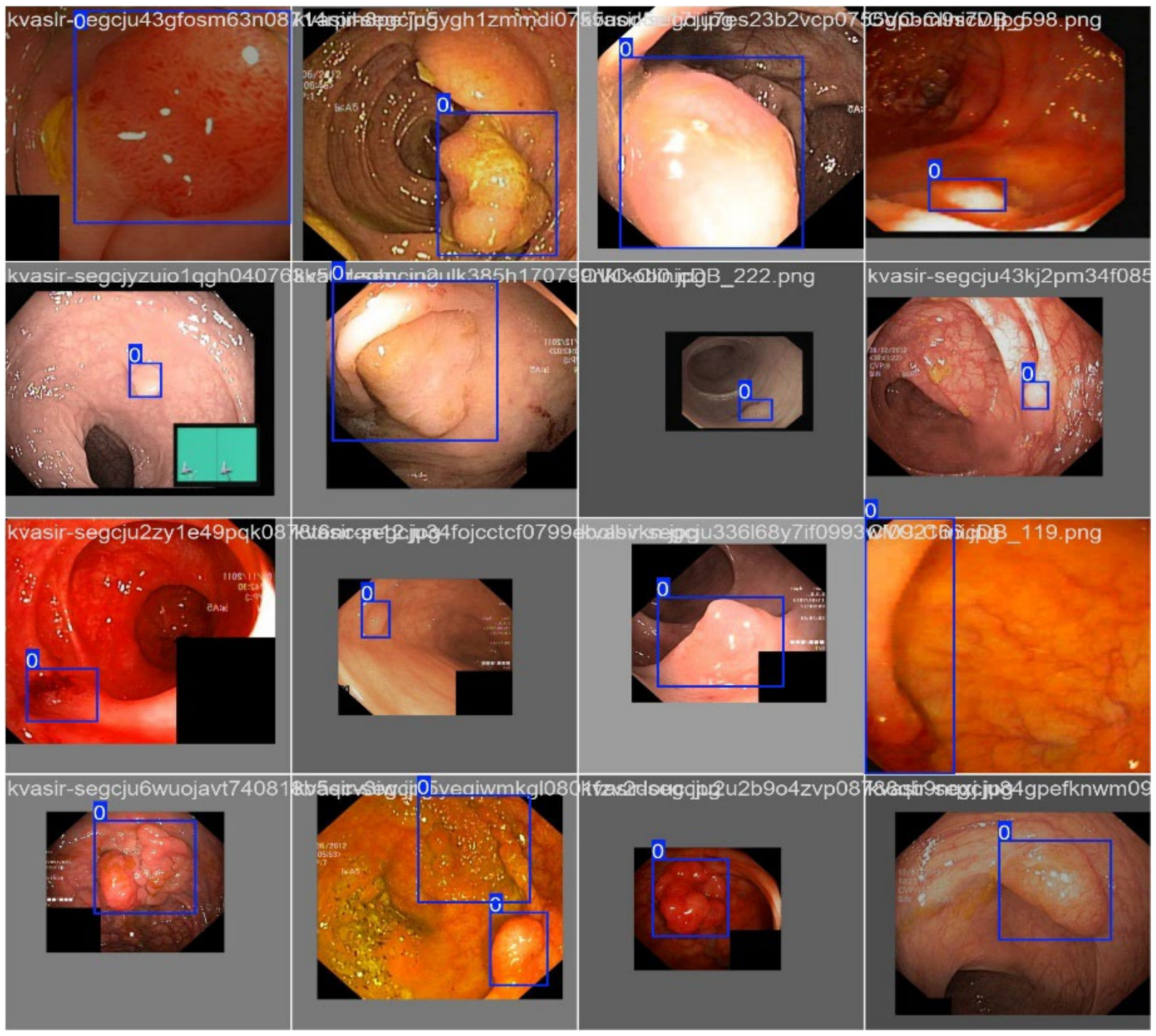

Figure 5. Sample images from Training Batch 8240 illustrating complex polyp cases.

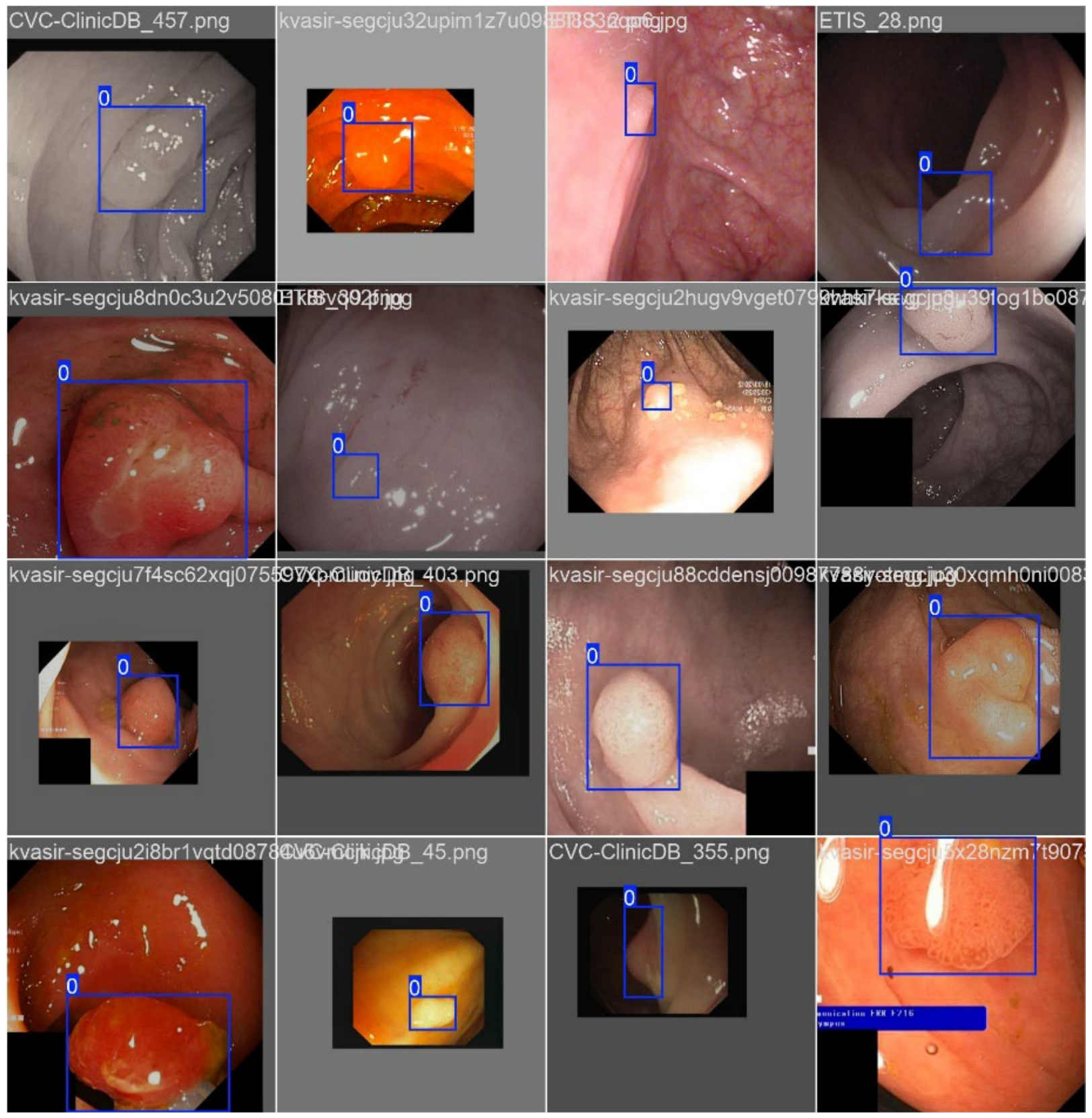

Figure 6. Sample images from Training Batch 8241 with large and irregular polyps.

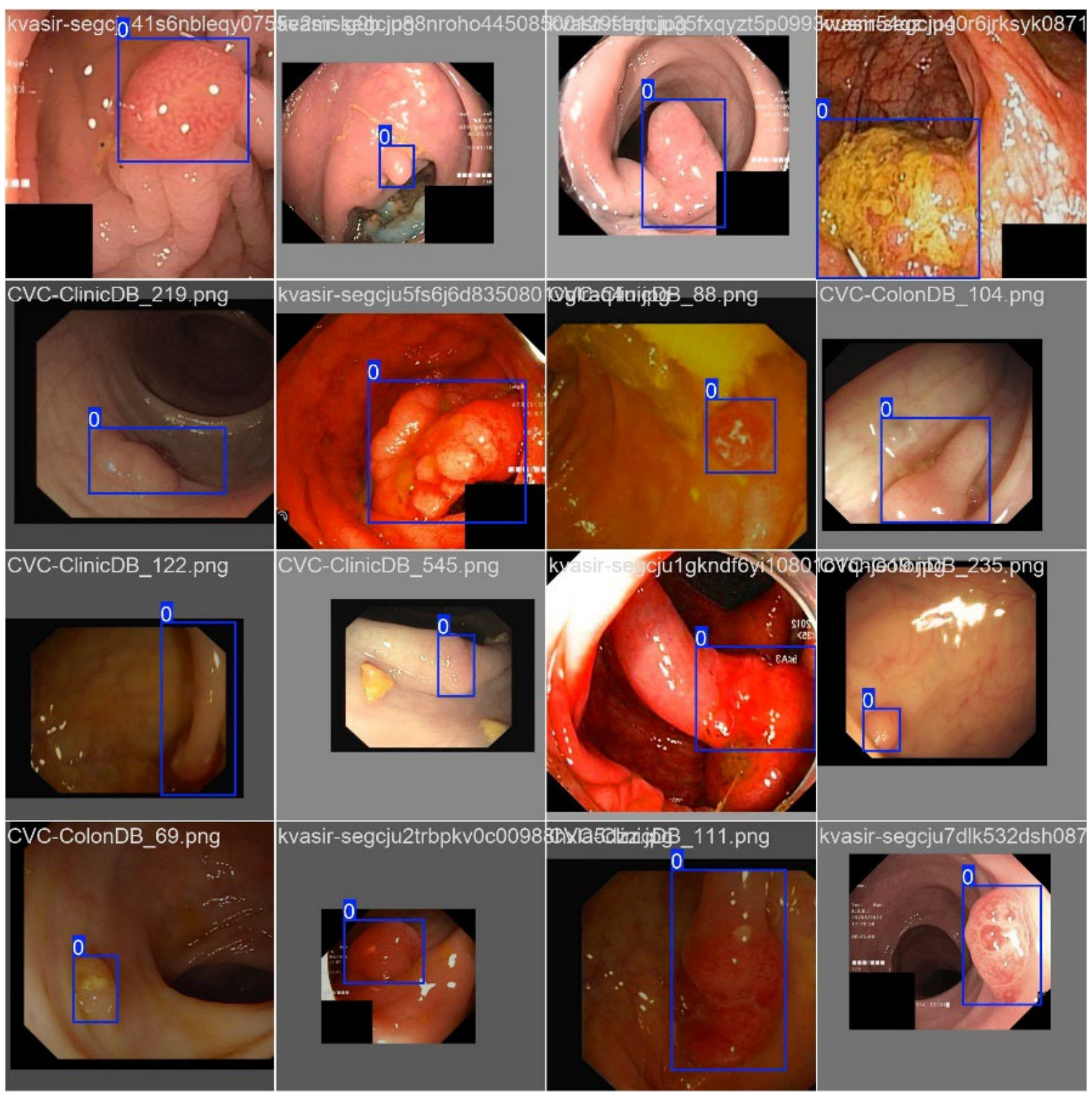

Figure 7. Sample images from Training Batch 8242 showing high-contrast polyps.

To better understand the distribution of annotated bounding boxes in our polyp detection dataset, we visualized key statistics using the images in Figures 8 and 9. Figure 8 presented an overview of the bounding box annotations. The top-left plot confirmed that all labeled instances belonged to the single class "polyp", while the top-right plot overlays all bounding boxes on a common frame, revealing that most polyps are located centrally within the image, a pattern consistent with typical endoscopic framing. The bottom plots showed heatmaps of the (x, y) center positions and the relationship between width and height. These indicated a central concentration of polyps and a strong correlation between width and height, suggesting polyps are generally circular or proportionally scaled.

Complementing this, Figure 9 provided a pairwise correlogram of bounding box parameters. The diagonal histograms illustrated that x and y were approximately normally distributed around 0.5, while width and height were skewed toward smaller values. The heatmaps below the diagonal highlighted a clear positive correlation between width and height, and suggested that centrally located polyps may be slightly larger. Together, these analyses confirmed that the dataset provided a realistic and varied representation of polyp appearances and positions, supporting robust model training.

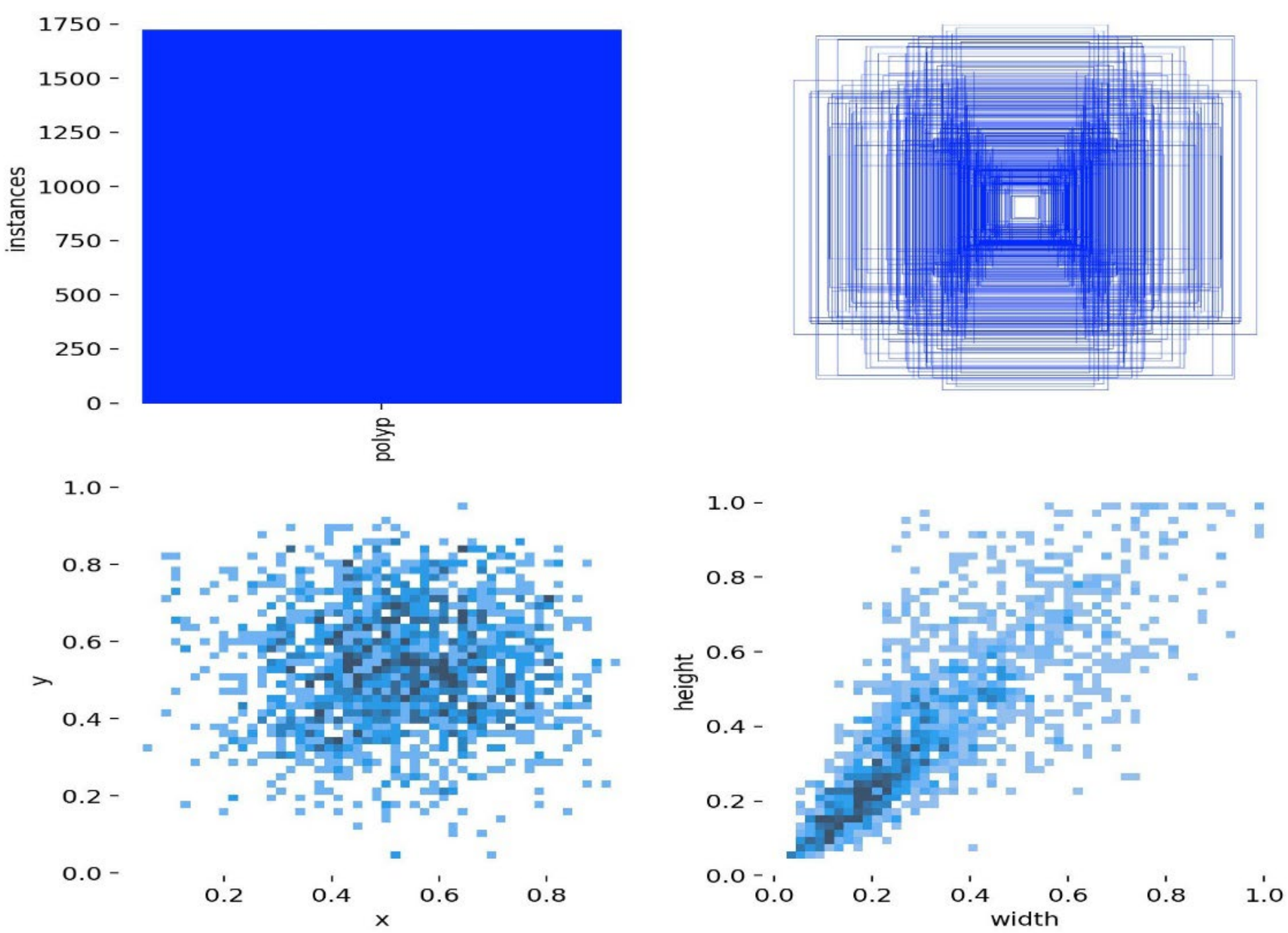

Figure 8. Distribution and Spatial Statistics of Polyp Bounding Boxes.

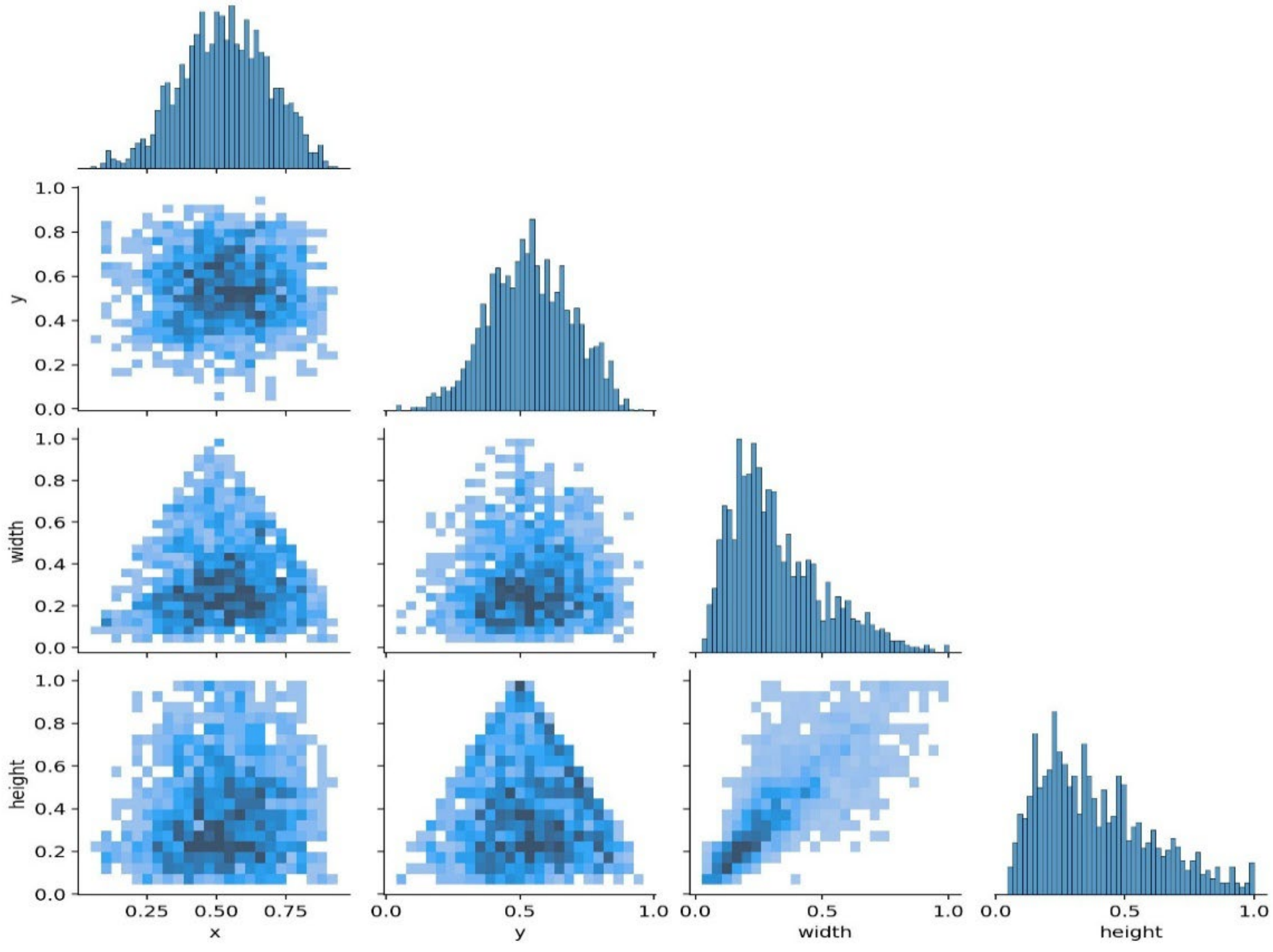

Figure 9. Pairwise Correlation Analysis of Bounding Box Parameters.

### 3.2. Outlier Detection Using LOF

To improve model performance by removing noisy data, an outlier detection step is incorporated using the LOF algorithm. LOF is a density-based, unsupervised anomaly detection technique. In this study, it is configured with 30 neighbors and a contamination rate of 5%. Data points with significantly lower density compared to their neighbors are identified and removed, thereby enhancing the quality of the training data.
The mathematical formulation of LOF is defined as follows (1):

$$LOF(p) = \frac{1}{|N_k(p)|} \sum_{t \epsilon N_k(p)} \frac{LRD_t}{LRD_p} \tag{1}$$

$$LRD_p = \frac{1}{\frac{1}{|N_k(p)|} \sum_{s \epsilon N_k(p)} reach_{dist_k}(p, s)}$$

$$reach_{dist_k}(p, s) = max\ (k - distance_r, d(p, r))$$

where $d(p, r)$ is the distance between the points *p* and *r, k_distance* of point *r* is the distance from *r to its* $k^{th}$ *nearest neighbor,* and $N_k(p)$ is a set of data points whose distance from point *p* is less than *k_distance*.
This helps remove data points with significantly low local density, which are likely to be anomalies.

### 3.3. YOLO-v11n Training Configuration

Following outlier removal, the YOLO-v11n model, a compact version of YOLO optimized by Ultralytics, is trained. This model balanced speed and accuracy, making it suitable for real-time medical image analysis. The training images are resized to 640x640 pixels, and the model is trained for 80 epochs with a batch size of 16. Table 2 listed the key hyperparameters used in the training process.
YOLO-v11n integrated advanced data augmentation techniques such as Mosaic, Mix-up, Copy-Paste, RandAugment, and Erasing. YOLO-v11 offered five scaled variants, and for this study, the nano version is chosen due to its balance of accuracy, speed, and minimal resource requirements.
YOLO-v11n maintained the fundamental architecture of earlier YOLO models, consisting of three main components: Backbone, Neck, and Head.

**1) Backbone**

Using convolutional neural networks to extract multi-scale feature maps from the input images and some other custom blocks, the Backbone performed mainly as the feature extraction component.
Initial CNN layers are utilized to increase the number of channels while down-sampling the input images. One of the most significant changes and improvements in the backbone of YOLOv11 is the replacement of the C2f block in its predecessors with the C3k2 block. This block utilized two smaller convolutional layers at the beginning and end of the block with smaller kernel sizes compared to one large convolutional layer, leading to a more efficient version of the Cross Stage Partial (CSP) Bottleneck. Spatial Pyramid Pooling Fast (SPPF) block from earlier versions is kept in YOLO-v11's architecture, but another block called Stage Partial with Spatial Attention (C2PSA) is used after it. This C2PSA block used the advantages of the self-attention mechanism and allowed the model to concentrate more precisely on significant areas in the image, resulting in higher detection accuracy, especially for objects that are smaller or partially occluded, and as a result, enhancing detection performance and accuracy for objects of different positions and sizes.

**2) Neck**

The neck architecture included upsampling and concatenating feature maps from multiple levels to merge multi-scale features and deliver them to the head to make predictions. This structure enabled the model to efficiently obtain multi-scale information. Another significant difference between YOLO-v11 and its predecessors was the replacement of the C2f by the C3k2. This replacement led to higher speed, efficiency, and performance.

### 3) Head

The responsibility of the head section was the generation of final predictions of bounding box regression and classification. Feature maps transmitted from the neck are processed, bounding box regression and classification are produced, and final detections were output here.
The main differences between Yolo-v11's head architecture and the previous versions were in the use of C3k2 and Convolution-BatchNorm-Silu (CBS) blocks. The CBS block is used after C3k2. These blocks extracted relevant features, normalized the data, and applied SiLU activation functions for refining the feature maps further and consequently enhancing object detection accuracy.
Table 2 summarized some important hyperparameters of YOLO-v11n that were used in the training process.

**Table 2.** Hyperparameters of the YOLO-v11n Model

| **Hyperparameter** | **Value** |
|---|---|
| Epochs | 80 |
| Batch Size | 16 |
| Image Size | 640 |
| Learning Rate (lr0) | 0.01 |
| Learning Rate Factor | 0.01 |
| Momentum | 0.937 |
| IOU Threshold | 0.7 |
| Max Detections | 300 |
| Mosaic Augmentation | 1 |

YOLO-v11n introduced architectural improvements such as the C3k2 block in the backbone and the C2PSA block for spatial attention, enhancing detection accuracy. The head used CBS (Convolution-BatchNorm-SiLU) blocks for efficient feature transformation. The architecture of YOLO-v11 is shown in Figure 10, and Table 3 illustrated a comparative analysis of the YOLO Nano model parameters.

**Table 3.** Comparison of YOLO Nano Model Parameters

| **Model Version** | **Model Type** | **Number of Parameters** |
|---|---|---|
| YOLO-v4 | Nano | 6.2 million |
| YOLO-v5 | Nano | 1.8 million |
| YOLO-v6 | Nano | 1.3 million |
| YOLO-v7 | Nano | 2.5 million |
| YOLO-v8 | Nano | 3.7 million |
| YOLO-v11 | Nano | 2.6 million |

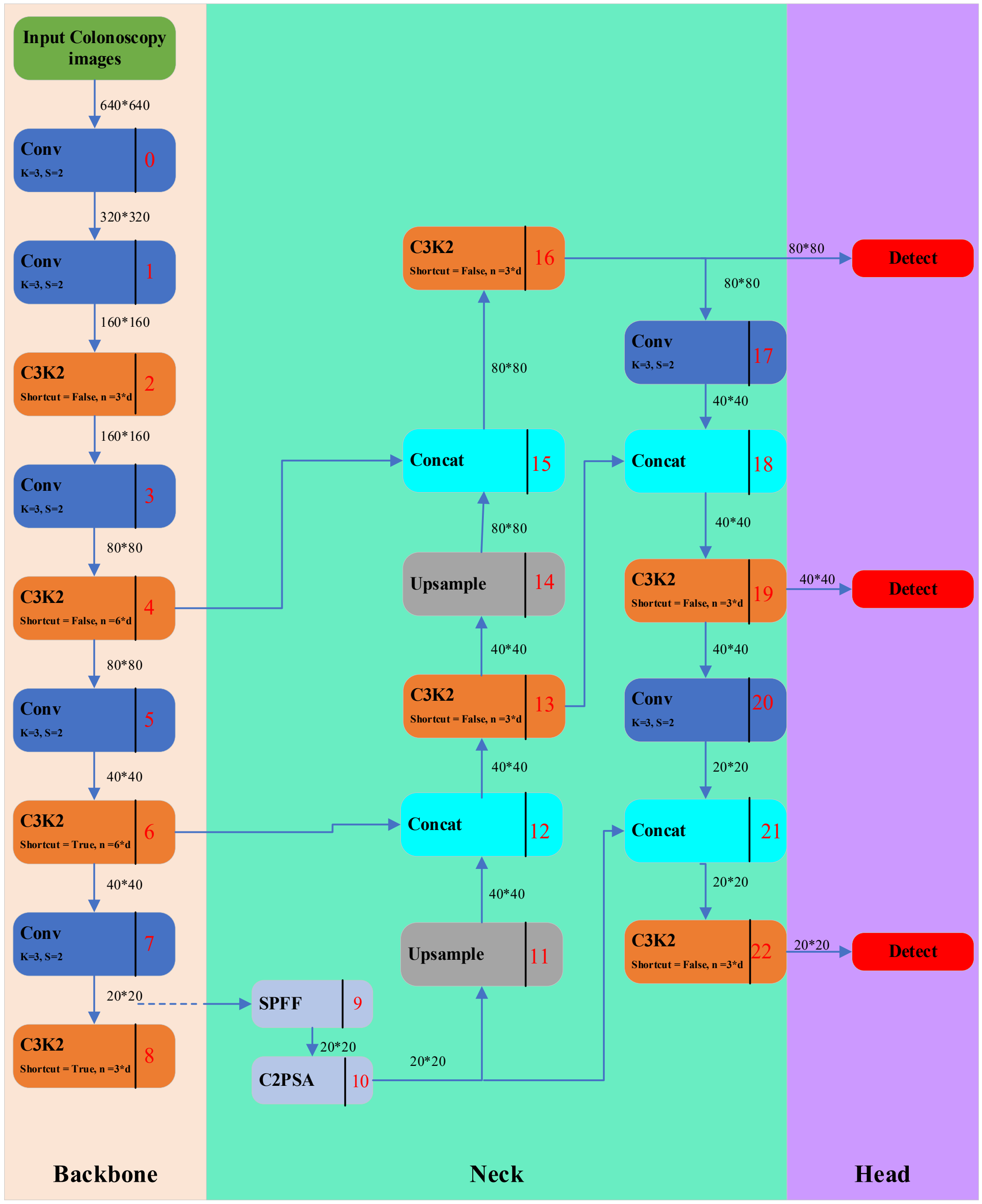

Figure 10. YOLO-v11 architecture, including Backbone, Neck, and Head.

### 3.4. Loss Function Design

In this article, we utilized a loss function of Multiple terms (Box_Loss, Cls_Loss, and Dfl_Loss) for Object Detection via YOLO-v11. These loss components worked together and enabled the model to precisely learn the bounding box localization and object classification at the same time. The total loss function applied in training is defined as (2):

$$Total_{Loss} = \lambda_{Box} Box_{Loss} + \lambda_{Cls} Cls_{Loss} + \lambda_{Dfl} Dfl_{Loss} \qquad (2)$$

Box_ loss played a significant role in the localization precision and directly impacts it. It calculated the difference between the ground truth and the predicted bounding box. It is defined by the equation below:

$$Box_{-Loss=1-CIoU=1-\left[IoU-\frac{\rho^2(b,b^2)}{c^2}-\alpha v\right]} \tag{3}$$

$$v = \frac{4}{\pi^2} \times \left( arctan\frac{\omega^2}{h^2} - arctan\frac{\omega}{h} \right)^2, \alpha = \frac{v}{(1 - IoU) + v}$$

where $IoU$ was the Intersection over Union between the ground truth and predicted bounding boxes, $\omega^2, h^2, \omega, h$ were the width and height of the predicted and ground truth bounding boxes, respectively. $\rho^2(b, b^2)$ was the Euclidean distance computed between the centers of the boxes, $c$ was the diagonal distance of the minimal enclosing bounding box, $v$ was the aspect ratio consistency term (penalized the difference between the ratio of ground truth and predicted boxes), and $\alpha$ was the adjustment factor.

Cls_Loss was an essential metric optimizing the performance of the model's object classification. It employed weighted binary cross-entropy (BCE) loss, as seen in the following formula:

$$Cls_{Loss} = -\sum_{j=1}^{M} \left[ \omega_j y_j log(p_j) + (1 - y_j) log(1 - p_j) \right] \tag{4}$$

Where $M$ was the total number of classifications, $y_j$ was the ground truth for sample $j$, $p_j$ was the predicted probability related to sample j, and $\omega_j$ was the weight.

For enhancing bounding box prediction, we utilized the Dfl_Loss (distributed focal loss), which concentrated on the accuracy of localization of the bounding box. It is defined as:

$$Dfl_{Loss} = \sum_{j=1}^{M} (p_j \times |x_j - x_j^2|) \tag{5}$$

where $x_j, x_j^2$ These were the coordinates of the bounding boxes for predicted and ground truth boxes.

## 4. Results

In this section, we presented performance results of the LOF + YOLOv11n model on the combined dataset. The first evaluated the overall performance of the proposed model, including its classification accuracy, detection reliability, error reduction during training, and generalization ability on validation data, and the second outlined the experimental setup of the implementation tools and parameter configurations. In Evaluation Metrics, we defined and justified the use of precision, recall, and F1-score for assessing segmentation performance. In the discussion section, we compared our model against recent approaches to highlight improvements. Finally, the Limitations subsection critically examined the model's constraints.

### 4.1 Evaluation

Figures 11 and 12 illustrated the confusion matrix and its normalized counterpart for one representative fold. In the raw confusion matrix in Figure 13, the model correctly identified most polyp instances (true positives = 275) while maintaining a relatively low number of false negatives (21) and false positives (37). The normalized matrix in Figure 11 further highlighted the performance balance, with a polyp detection rate of 93% and a near-perfect background classification rate of 100%.

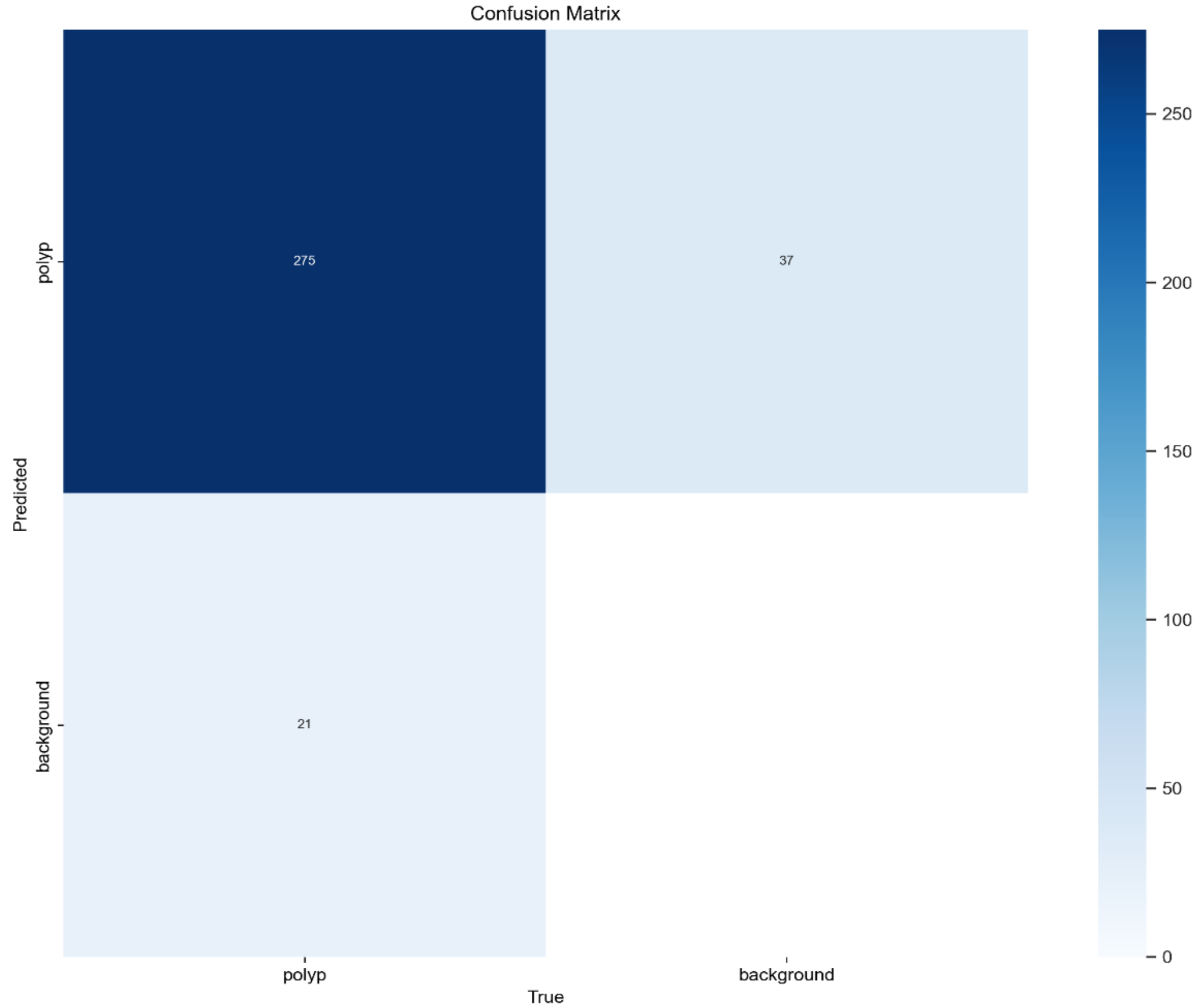

Figure 11. Confusion Matrix Showing Classification Results for Polyp and Background Classes.

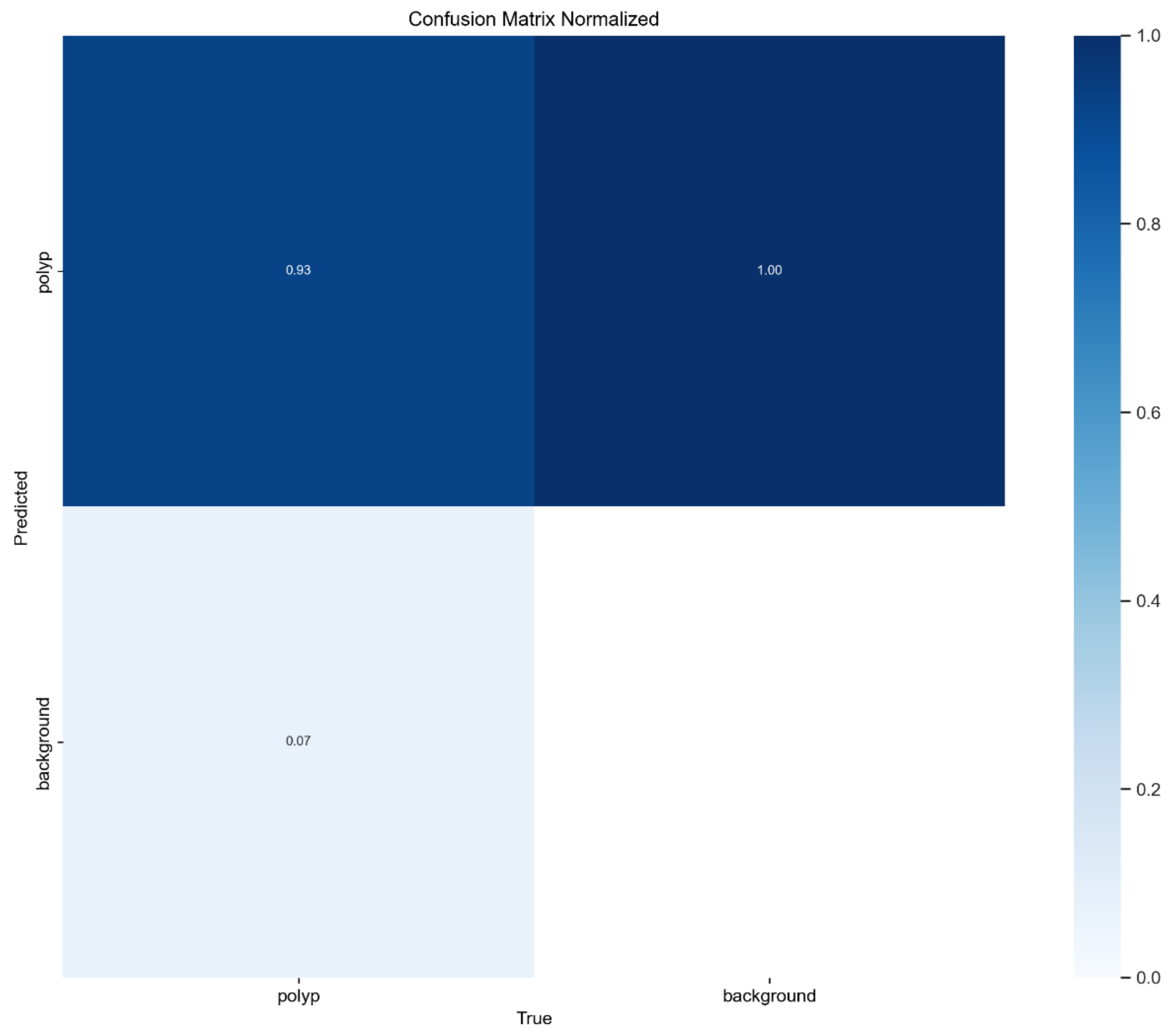

Figure 12. Normalized Confusion Matrix of Detection Model Performance on Polyp Classification.

We reported confidence-based curves for F1-score, Precision, and Recall. These plots were generated from the evaluation folder corresponding to the trained models.

The F1-confidence curve in Figure 13 illustrated the harmonic meaning between Precision and Recall across varying confidence thresholds. The model achieved its optimal F1-score of 0.91 at a confidence threshold of 0.357, indicating a strong balance between minimizing false positives and false negatives. This stability across a wide confidence range reflected the robustness of the LOF-based preprocessing step, which effectively reduced noise and enhanced discriminative feature learning for YOLO-v11n. The curve’s plateau around high F1 values confirmed that the framework consistently maintained high detection accuracy irrespective of slight variations in thresholding.

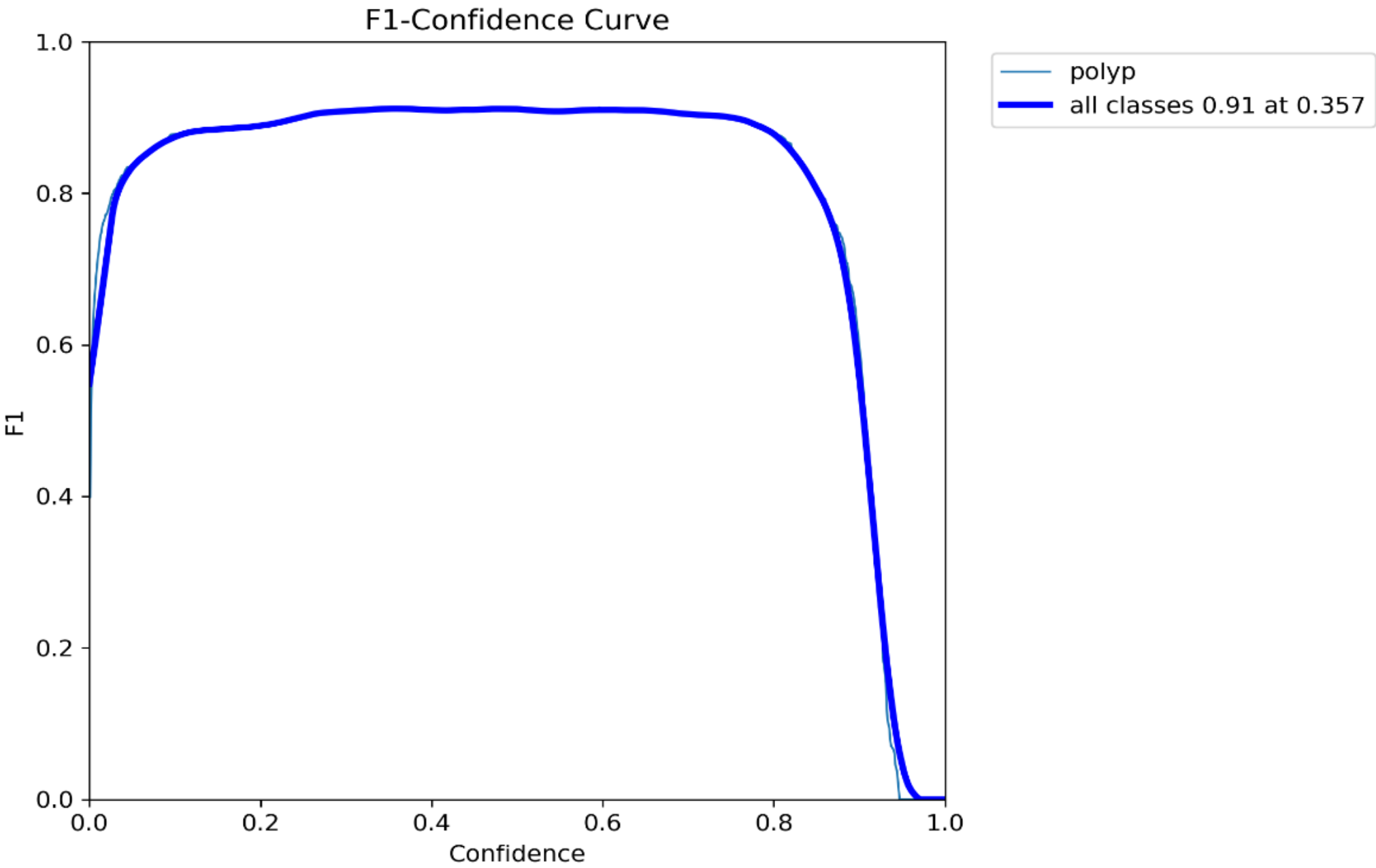

Figure 13. F1-Confidence Curve of the Proposed Model.

The Precision-confidence curve in Figure 14 demonstrated a steady increase in Precision with higher confidence thresholds, ultimately reaching 1.0 at a threshold of 0.933. This suggested that when the model assigns higher confidence to its detections, they are almost always correct. Such behavior was critical for clinical deployment, where minimizing false alarms was paramount. The trend also highlighted the lightweight model's ability to effectively suppress false positives, further validating the integration of LOF preprocessing for noise reduction.

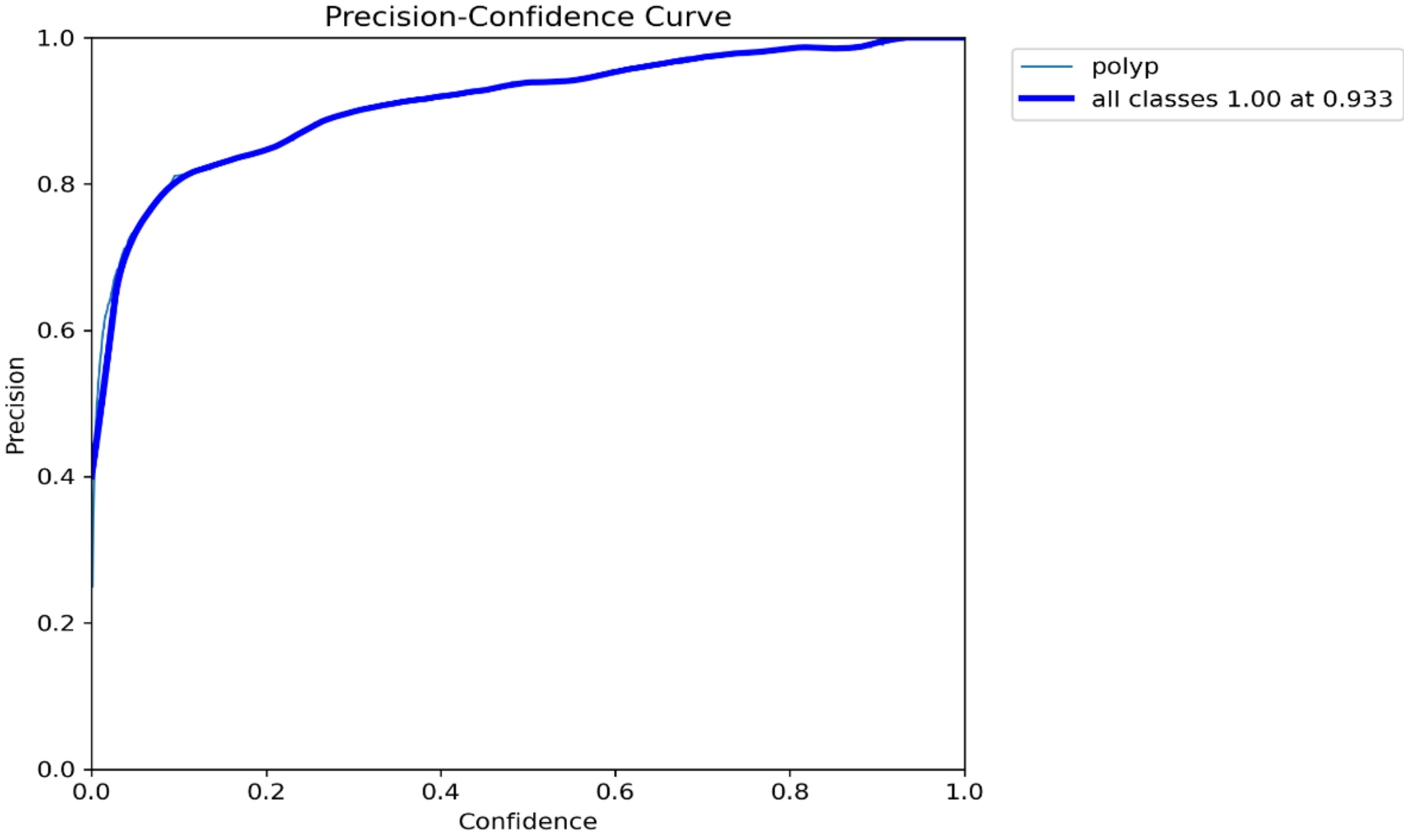

Figure 14. Precision-Confidence Curve of the Proposed Model.

The Recall-confidence curve in Figure 15 started at a high value of 0.99 at the lowest threshold and gradually decreased as the confidence threshold increased. This was consistent with the expected trade-off between Recall and Precision: while lower thresholds ensured that nearly all polyps are detected, stricter thresholds filter out uncertain predictions, thereby reducing Recall. Importantly, the relatively slow decline until around 0.8 confidence demonstrated the framework's robustness in capturing true positives even at moderately strict decision boundaries.

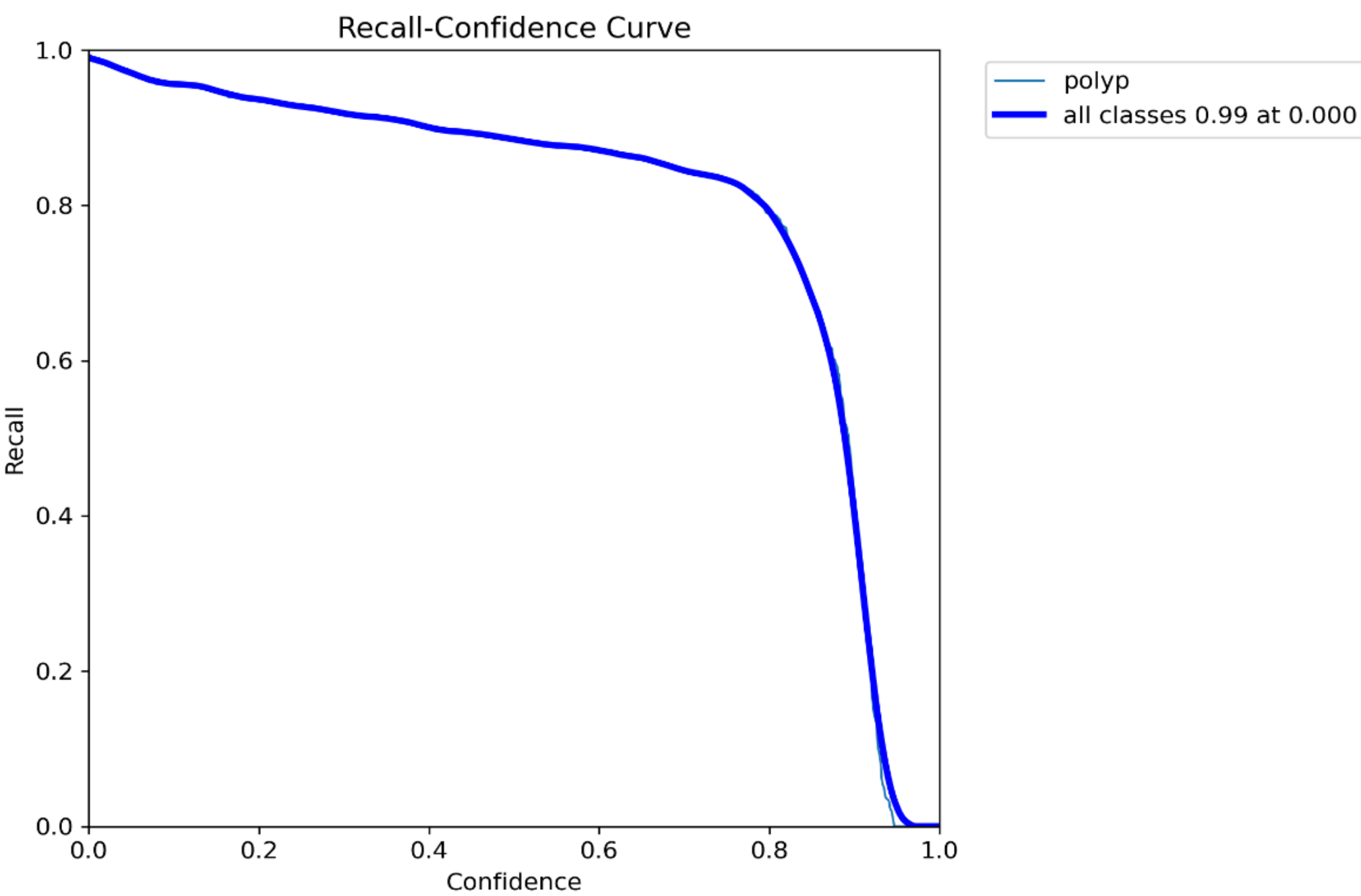

Figure 15. Recall-Confidence Curve of the Proposed Model.

The PR curve in Figure 16 illustrated the trade-off between precision and recall across different classification thresholds. In the presented curve, we observed a high precision across a broad range of recall values, which was indicative of a well-performing object detection model. The area under the PR curve (mAP@0.5) for the "polyp" class reached 0.965, which reflected the model's ability to accurately detected polyps with minimal false positives or false negatives.

The blue line represented the aggregate performance across all folds and corresponded to **(**mAP**)** at an IoU threshold of 0.5. This high mAP score confirmed that the model maintains consistent detection accuracy during all five validation rounds, thereby reinforcing its robustness and generalizability. This level of performance is particularly significant given the constraints of real-time clinical deployment, where detection speed and accuracy must be balanced. The integration of LOF**-based** preprocessing likely contributed to enhanced robustness by suppressing outlier noise in the input images, thereby improving the precision-recall behavior of the YOLOv11n detector.

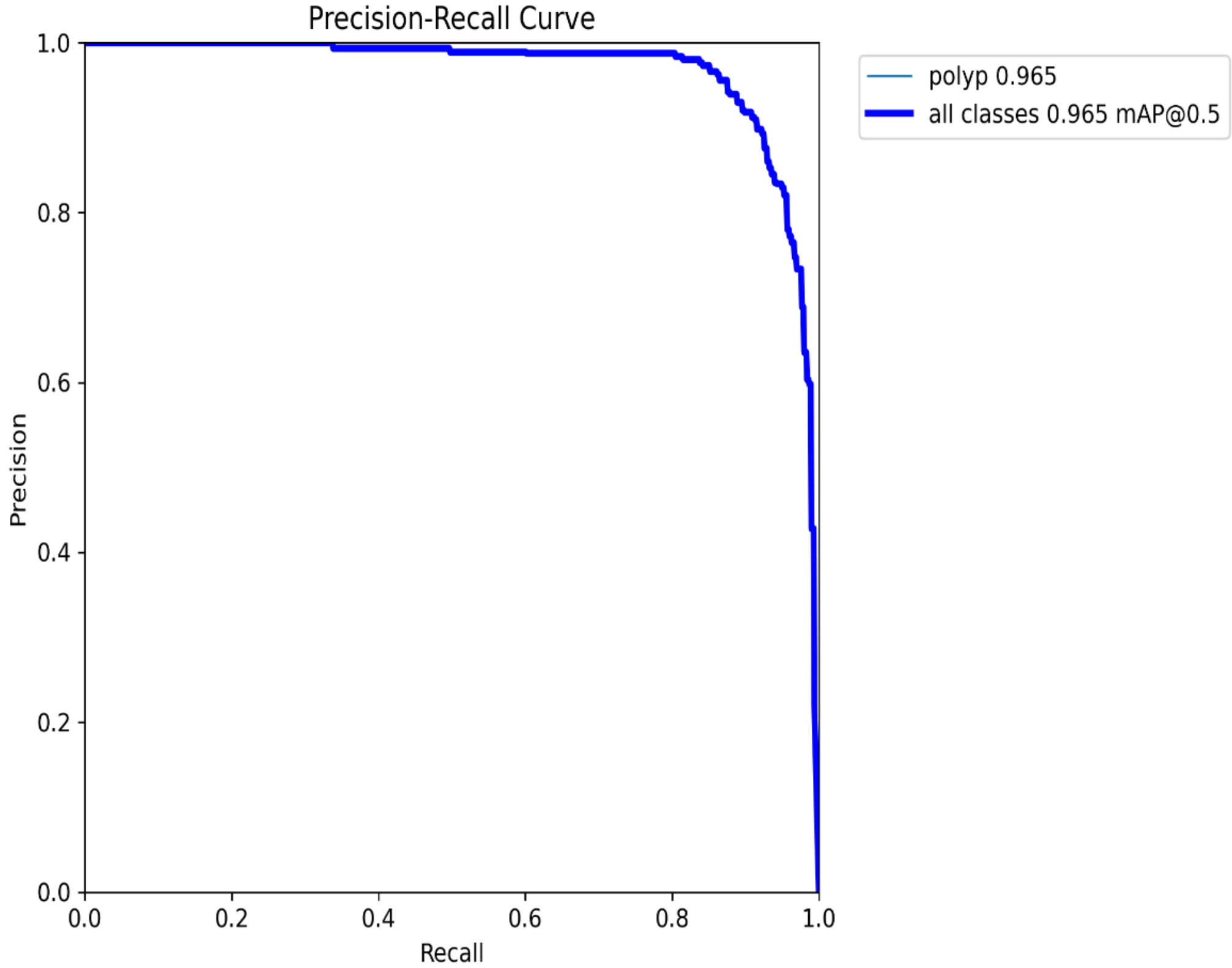

Figure 16. Precision-Recall Curve of the Proposed Model.

The results in Figure 17 underscore the effectiveness of the proposed polyp detection system. The low and steadily decreasing loss values across both training and validation indicated stable optimization and absence of overfitting. Additionally, the high precision, recall, and mAP scores confirmed the suitability of the proposed lightweight YOLOv11n architecture augmented with LOF-based preprocessing for real-time and accurate colorectal polyp detection.

Figure 17 presented a comprehensive visualization of the training and validation performance metrics for our proposed YOLOv11n-based framework, integrated with LOF-based preprocessing, evaluated under a 5-fold cross-validation setup. The first three plots in the top row (train/box_loss, train/cls_loss, train/dfl_loss) represented the evolution of the bounding box regression loss, classification loss, and distribution focal loss (DFL) during training. Corresponding plots in the bottom row (val/box_loss, val/cls_loss, val/dfl_loss) showed the same metrics during validation. Across all metrics, a consistent downward trend is observed over the course of ~80 epochs, indicating effective convergence of the model. The training and validation losses exhibit closely matched curves, suggesting minimal overfitting and strong generalization ability.

The plots labeled metrics/precision(B) and metrics/recall(B) reflected the evolution of the model's precision and recall on the validation set across epochs. Both metrics improved steadily, with final values exceeding 0.9, indicating high reliability in both detecting and correctly classifying polyps. The metrics/mAP50(B) and metrics/mAP50-95(B) plots depict the mean Average Precision (mAP) at IoU thresholds of 0.5 and the more stringent 0.5–0.95 range, respectively. These metrics also exhibit consistent improvement, converging to approximately 0.96 (mAP@0.5) and 0.78 (mAP@0.5–0.95).

The combination of fast convergence, strong generalization, and high detection accuracy makes this framework a compelling candidate for deployment in clinical endoscopic environments.

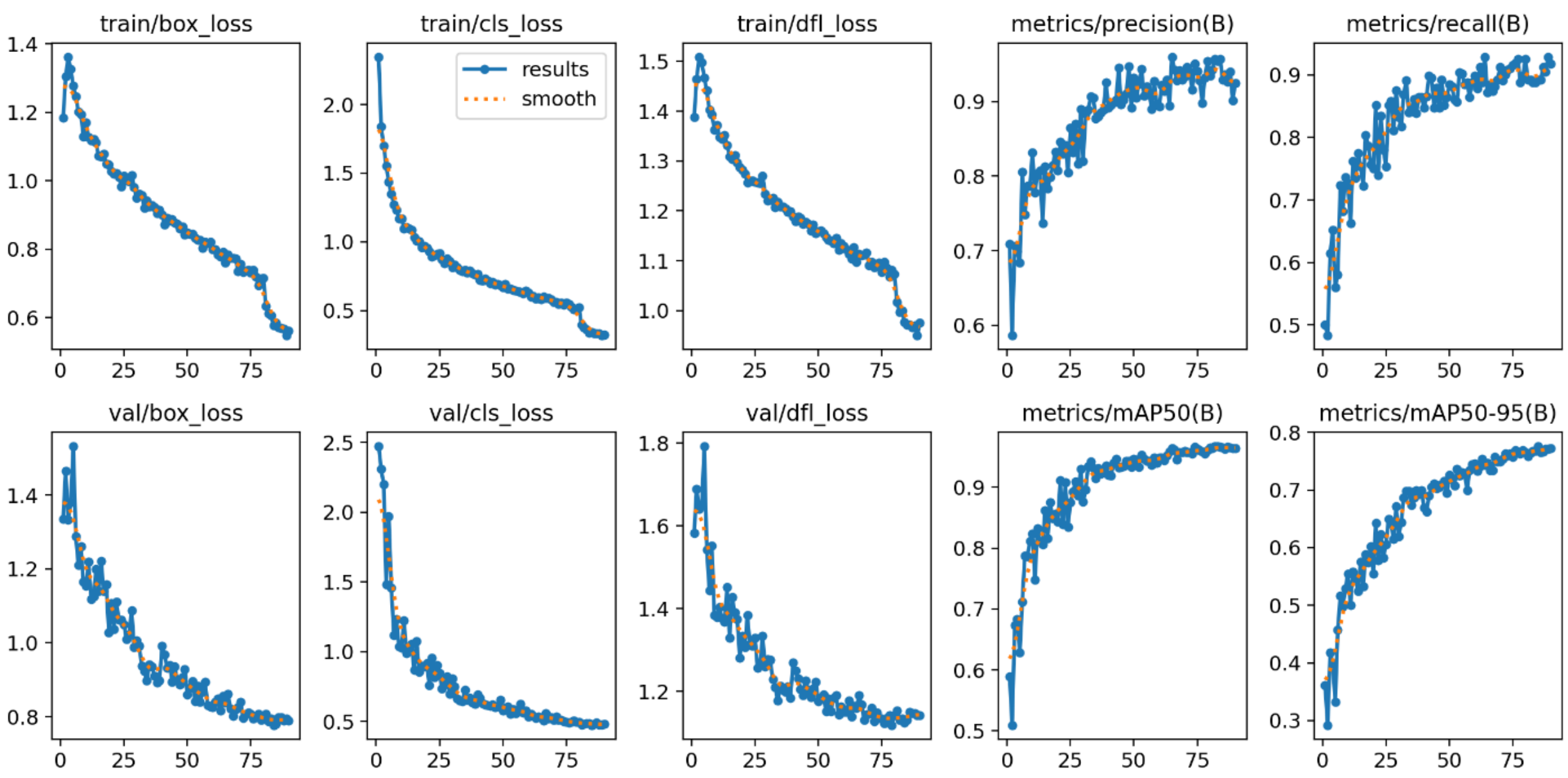

Figure 17. Training and Validation Metrics Across Epochs.

### 4.2. Experimental setup

The following configuration was used for efficient model training and evaluation.

#### 4.1.1. System configuration

All experiments were conducted on a machine equipped with an 11th Generation Intel® Core™ i7-11800H processor operating at 2.30 GHz, supported by 16 GB of RAM. The system ran on a 64-bit version of Windows 11 and utilized a CUDA-enabled NVIDIA GeForce RTX 3050 Ti GPU for accelerated deep learning computations. The development environment included Python 3.x as the programming language, with Visual Studio Code (VS Code) serving as the Integrated Development Environment (IDE). The system type was 64-bit, x64-based architecture, and the device did not support pen or touch input.

#### 4.1.2 Training framework and experimental procedure

The experiments were implemented using the Ultralytics YOLO framework, specifically the YOLO11n model version (yolo11n.pt). Model training and evaluation were performed using a 5-fold cross-validation strategy to ensure robustness and generalization of results. For anomaly detection, the isolation forest algorithm from the scikit-learn library was employed. This approach allowed the integration of both object detection and unsupervised anomaly identification.

#### 4.1.3 Python libraries used

Several Python libraries, as shown in Table 4, were used to support the experimental workflow.

**Table 4.** Comparison of YOLO Nano Model Parameters

| Library | Purpose |
|---|---|
| **Os** | File path management and directory handling |
| **Numpy** | Numerical computations and array operations |
| **PIL (Pillow)** | Image loading and preprocessing |
| **sklearn. ensemble** | Isolation forest for anomaly detection |
| **Ultralytics** | Model training and evaluation using the YOLO framework |
| **Torch (PyTorch)** | Backend deep learning operations for YOLO |
| **opencv-python** | Image and video preprocessing (used as needed) |
| **yaml / ruamel.yaml** | Configuration file reading/editing (e.g., data.yaml) |

### 4.2 Evaluation metrics

To comprehensively assess model performance, the following evaluation metrics were employed: Precision, Recall, F1-score, Accuracy, mean Average Precision at IoU threshold 0.5 (mAP@0.5), and the more comprehensive mean Average Precision across multiple IoU thresholds (mAP@0.5:0.95). The formulas for these indices are as follows:

$$Precision = TP / (TP + FP) \quad (6)$$

$$Recall = TP / (TP + FN) \quad (7)$$

$$F1\text{-}Score = 2 * (precision * recall) / (precision + recall) \quad (8)$$

• True positives (TP): number of polyp items correctly predicted as polyps.
• True negatives (TN): number of non-polyp items correctly predicted as non-polyp.
• False positives (FP): number of non-polyp items incorrectly predicted as polyps.
• False negatives (FN): number of polyp items incorrectly predicted as non-polyp.

Precision in the context of polyp detection indicated the confidence level when a positive detection is made, and Higher precision reduced false alarms and financial stress. Recall, which is the fraction of detected objects, is crucial for timely detection and treatment, reducing mortality and preventing additional costs. Both metrics are essential in polyp detection. The F1-score provided a balanced assessment of both.

## 5. Discussion

In this study, several detection algorithms were implemented and systematically compared to evaluate their effectiveness in colorectal polyp detection. The architectures considered include YOLOv8, YOLOv9, YOLOv10, YOLOv11n, and the customized LOF-YOLOv11n. These models represented successive advancements in the YOLO family, each incorporating architectural refinements aimed at improving detection accuracy, computational efficiency, and robustness in challenging medical imaging scenarios. This comparative analysis is driven by recent studies highlighting the importance of adapting deep learning models to the unique challenges of endoscopic imaging, including issues like low contrast, inconsistent lighting, and complex, noisy backgrounds.
The comparative results, summarized in Table 5, demonstrate that the proposed LOF-YOLOv11n achieved superior performance across most key evaluation metrics. Specifically, it obtained the highest precision (95.83%), recall (91.85%), and F1-score (93.48%), while also achieving a leading mAP@0.5 score of 96.48%. Although YOLOv10 reported the highest mAP@0.5:0.95 (78.81%), its recall (88.46%) was notably lower, suggesting a performance trade-off between strict localization accuracy and the ability to detect all true positives. By contrast, LOF-YOLOv11n provided a more balanced outcome, improving sensitivity without compromising precision. This is of particular clinical importance, as minimizing false negatives directly reduced the likelihood of missed polyps, a key factor in colorectal cancer prevention.

The improvements achieved by LOF-YOLOv11n can be attributed to the integration of (LOF) in preprocessing, which effectively reduced the impact of noisy or low-contrast regions. This enhancement aligned with findings in prior research, where robust preprocessing combined with a lightweight detection architecture has been shown to yield performance in medical imaging applications. The observed results confirm that while baseline YOLO variants such as YOLOv8 and YOLOv9 delivered competitive accuracy, domain-specific customization remains essential for optimal performance in polyp detection tasks.

LOF-YOLOv11n demonstrated the best overall trade-off among the tested models, combining high precision, superior recall, and competitive mAP scores. These outcomes highlight its potential for real-world deployment in colonoscopy workflows, where real-time efficiency and diagnostic reliability are both crucial.

**Table 5.** Comparison of YOLO

| Method | Precision | Recall | F1-Score | mAP@0.5 | mAP@0.5:0.95 |
|---|---|---|---|---|---|
| LOF-yolov11n | **95.83** | **91.85** | **93.48** | **96.48** | **77.75** |
| Yolov11n | **94.33** | **91.26** | **92.75** | **96.25** | **77.57** |
| yolov10 | **94.28** | **88.46** | **91.27** | **95.16** | **78.81** |
| yolov9n | **94.05** | **91.09** | **92.94** | **95.99** | **76.44** |
| Yolov8n | **94.21** | **91.65** | **91.85** | **95.34** | **76.25** |
| Yolov8s | **94.51** | **91.26** | **92.36** | **95.7** | **76.57** |

Figure 18 presented a comparative evaluation of several lightweight YOLO variants (YOLOv8n/s, YOLOv9n, YOLOv10, YOLOv11n) against our proposed LOF-YOLOv11n model across five widely adopted object detection metrics: Precision, Recall, F1-Score, mAP@0.5, and mAP@0.5:0.95.

- Precision: LOF-YOLOv11n consistently outperformed the baselines, showing ~0.5–1% improvement over YOLOv11n and other counterparts. This indicated a reduction in false positives, which is consistent with the expected benefit of integrating outlier-aware filtering (LOF) in post-processing.
- Recall: While all models maintain high recall, YOLOv8n exhibited a noticeable drop, attributable to its smaller parameter budget. Importantly, LOF-YOLOv11n preserved recall at a high level, suggesting that the additional filtering did not sacrifice true positive detections.
- F1-Score: With increased precision and stable recall, LOF-YOLOv11n achieved the best balance between sensitivity and specificity, reflected in the highest F1 scores among all tested models.
- mAP@0.5: The proposed model yielded the highest or near-highest performance, outperforming other versions by approximately 0.3–1 percentage point, confirming its effectiveness in joint classification and localization at the IoU threshold of 0.5.
- mAP@0.5:0.95: All models demonstrated the expected drop in performance when evaluated over stricter IoU thresholds, with a ~18–20% gap relative to mAP@0.5. Nonetheless, LOF-YOLOv11n remained competitive, highlighting its robustness across varying localization requirements.

From an architectural perspective, the progression from YOLOv8 → YOLOv9n → YOLOv10 → YOLOv11n showed steady improvements in accuracy and stability, with larger variants (e.g., YOLOv8s) outperforming “nano” versions at the cost of efficiency. LOF-YOLOv11n introduced consistent gains across Precision, F1, and mAP@0.5, which can be attributed to more effective suppression of noisy or spurious predictions, without negatively impacting recall.

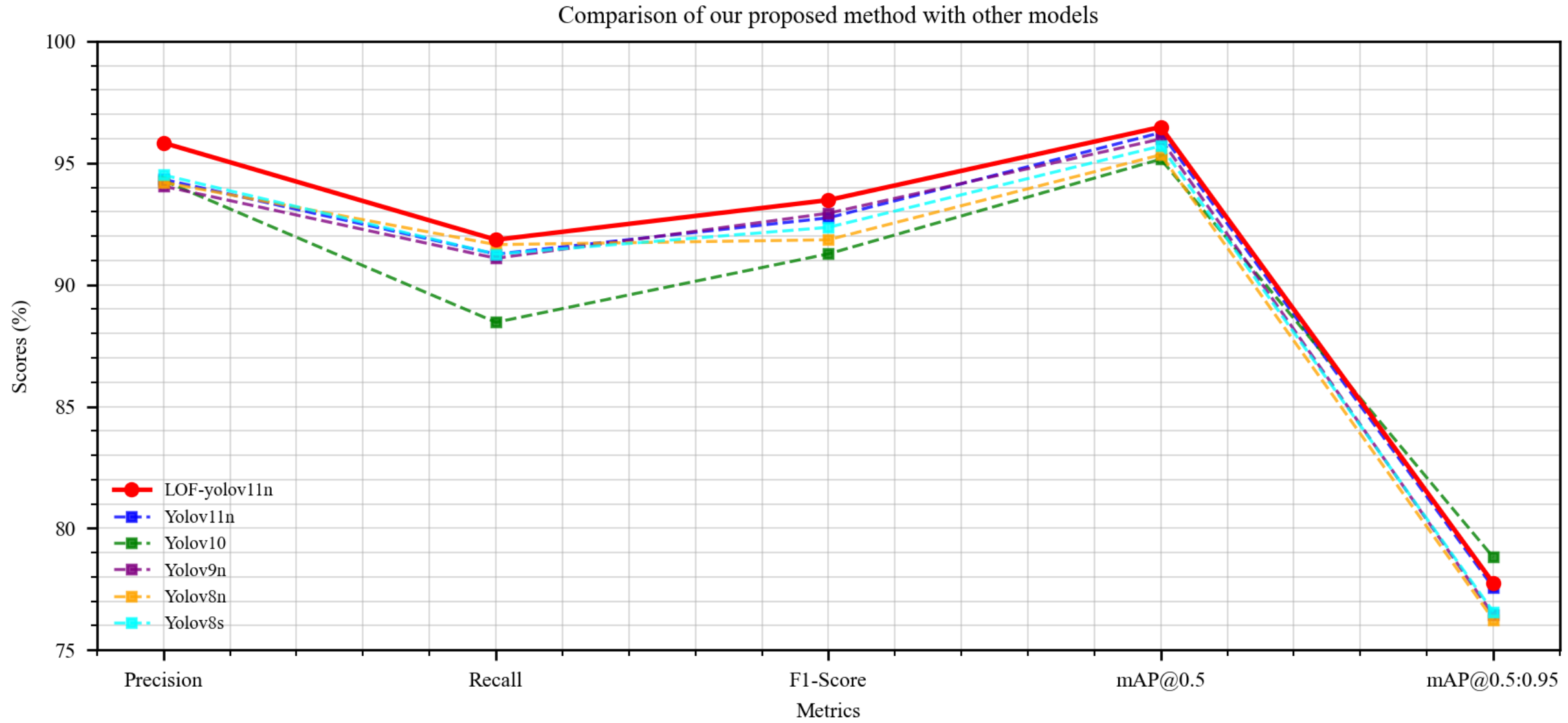

Figure 18. Line Chart Comparison of Proposed Method with YOLO Variants Based on Key Evaluation Metrics.

Figure 19 showed that LOF-YOLOv11n consistently achieved higher precision and F1-score, indicating an improved balance between sensitivity and specificity. In terms of mAP@0.5, the proposed model outperformed all baselines, confirming its effectiveness in joint classification and localization tasks. Although performance drops are observed for all models under the stricter mAP@0.5:0.95 metric, LOF-YOLOv11n maintained competitive results, demonstrating robustness across varying IoU thresholds.

Overall, Figure 19 highlighted that the integration of LOF enhanced detection reliability by reducing false positives without compromising recall, making the model particularly suitable for precision-critical applications such as surveillance and industrial inspection.

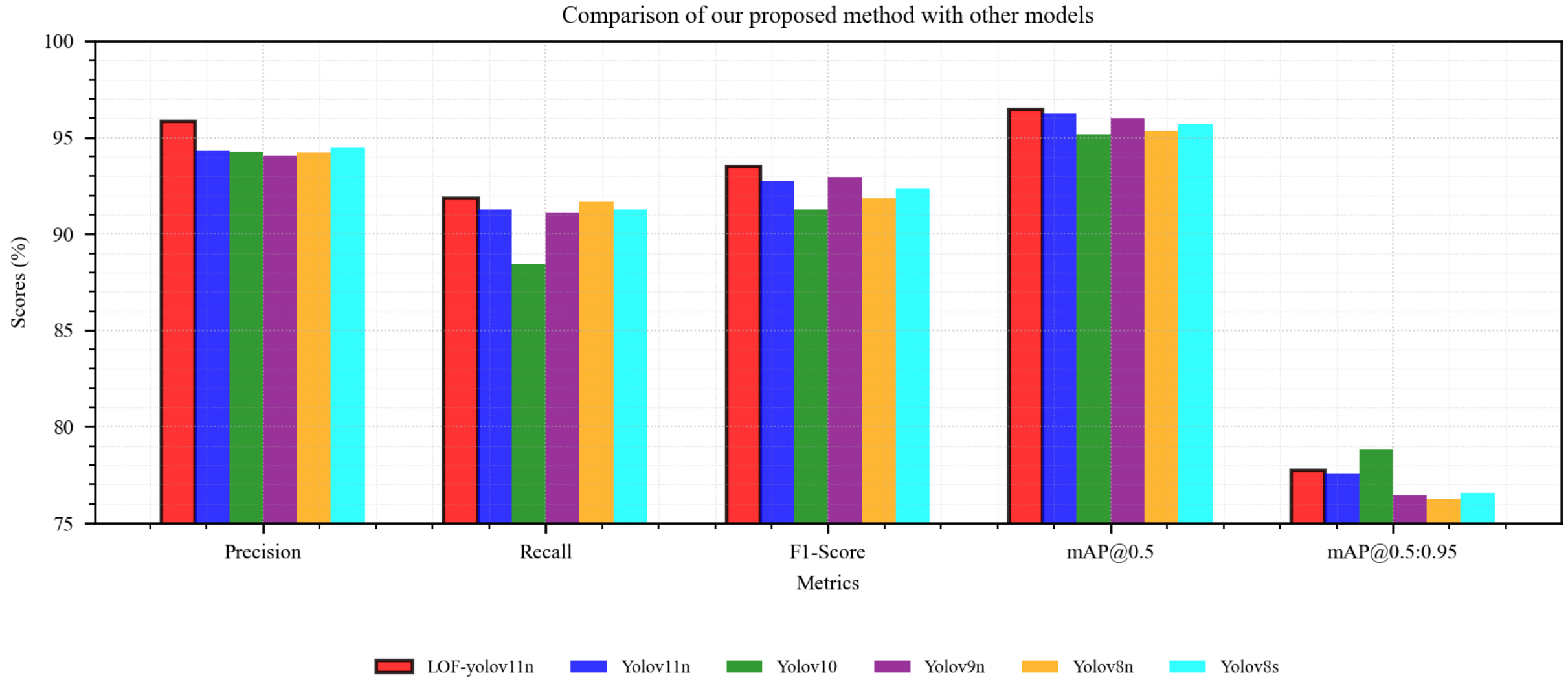

Figure 19. Bar Chart Comparison of LOF-YOLOv11n and Other YOLO Models Across Precision, Recall, F1-Score, and mAP.

Cross-validation is a widely accepted approach in medical image analysis because it allows for a more reliable estimation of generalization performance by training and validating multiple disjoint subsets of the data. This ensures that the reported metrics are not biased toward a specific data split and enhances the statistical significance of the results.

To visually assess the detection capability of our proposed system, we compared ground truth annotations with the model's predictions on three representative validation batches. For each batch, the upside images corresponded to the ground truth labels, while the bottom-side images showed the model predictions with confidence scores.

Images 20 and 21 in batch 0 illustrate that the ground truth labels clearly mark the polyp regions using blue bounding boxes. In the corresponding predictions, the model accurately localizes and classifies the majority of polyps with high confidence values **(~0.9)**. The bounding boxes in the predictions closely match the ground truth in both size and position, demonstrating precise spatial localization. The model shows strong robustness across varying lighting conditions and different anatomical appearances of polyps.

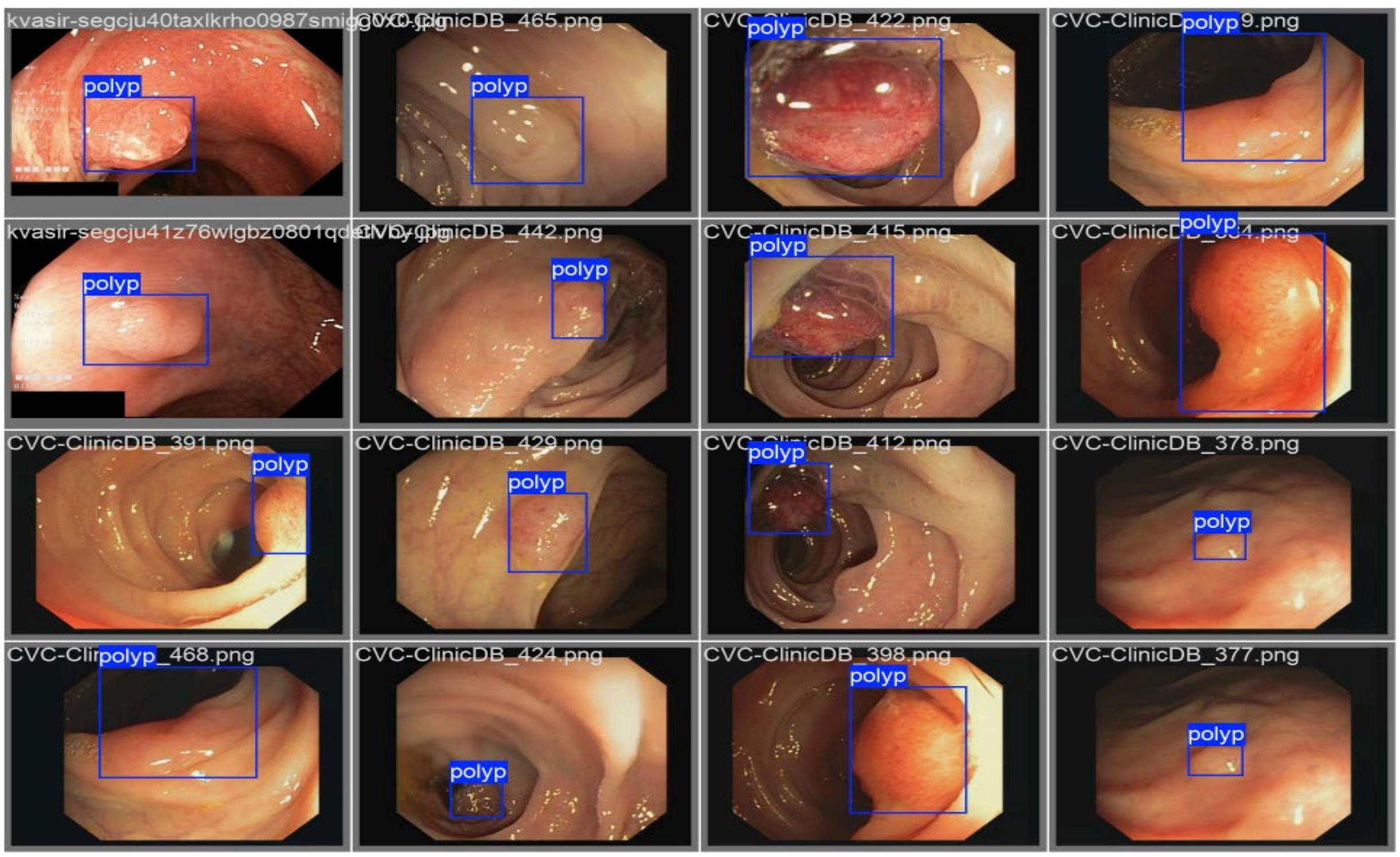

Figure 11. Ground truth bounding boxes indicating the location of polyps in colonoscopy images in batch 0.

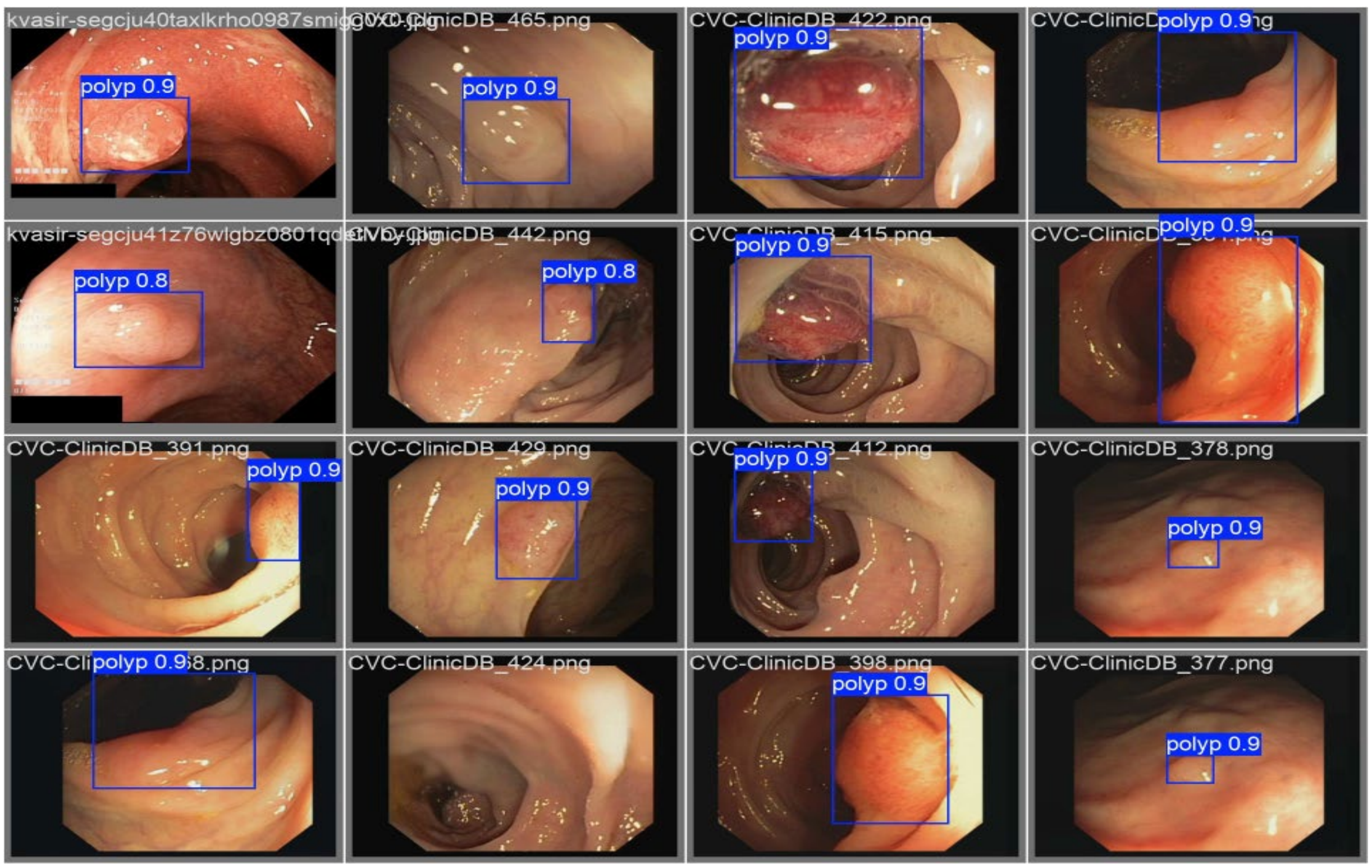

Figure 12. Predicted bounding boxes with confidence scores produced by the deep learning model in batch 0.

Images 22 and 23 in batch 1 illustrated that more visually complex samples, with some polyps partially occluded or embedded within mucosal textures. Despite these challenges, the model maintained high sensitivity, detecting nearly all labeled polyps with minimal deviation in bounding box positioning. Confidence scores remained consistent (~0.8–0.9), indicating the model's confidence in diverse visual contexts. The predictions maintain alignment with the ground truth, affirming the generalizability of the framework across different patients and image variations.

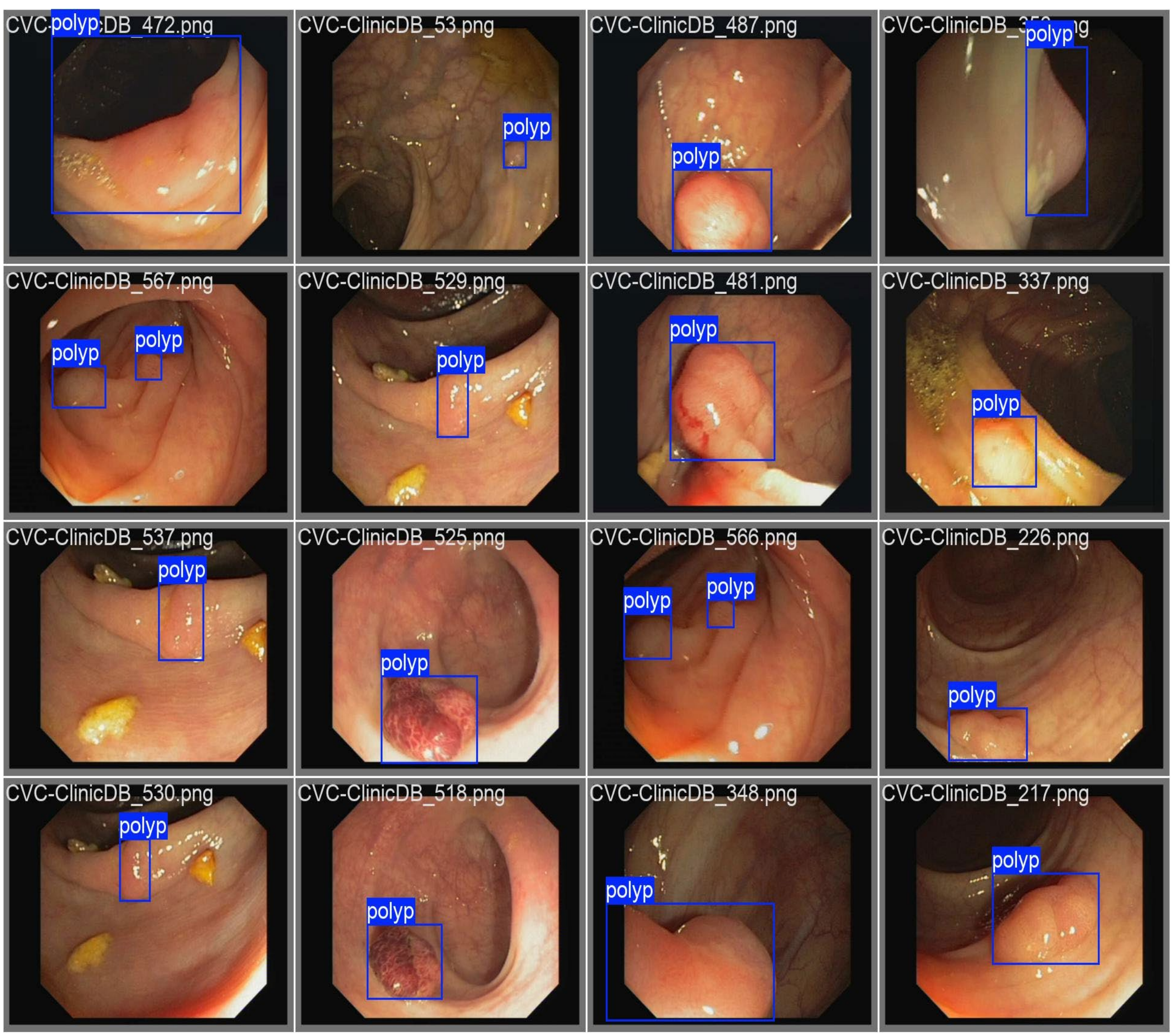

Figure 13. Ground truth bounding boxes indicating the location of polyps in colonoscopy images in batch 1.

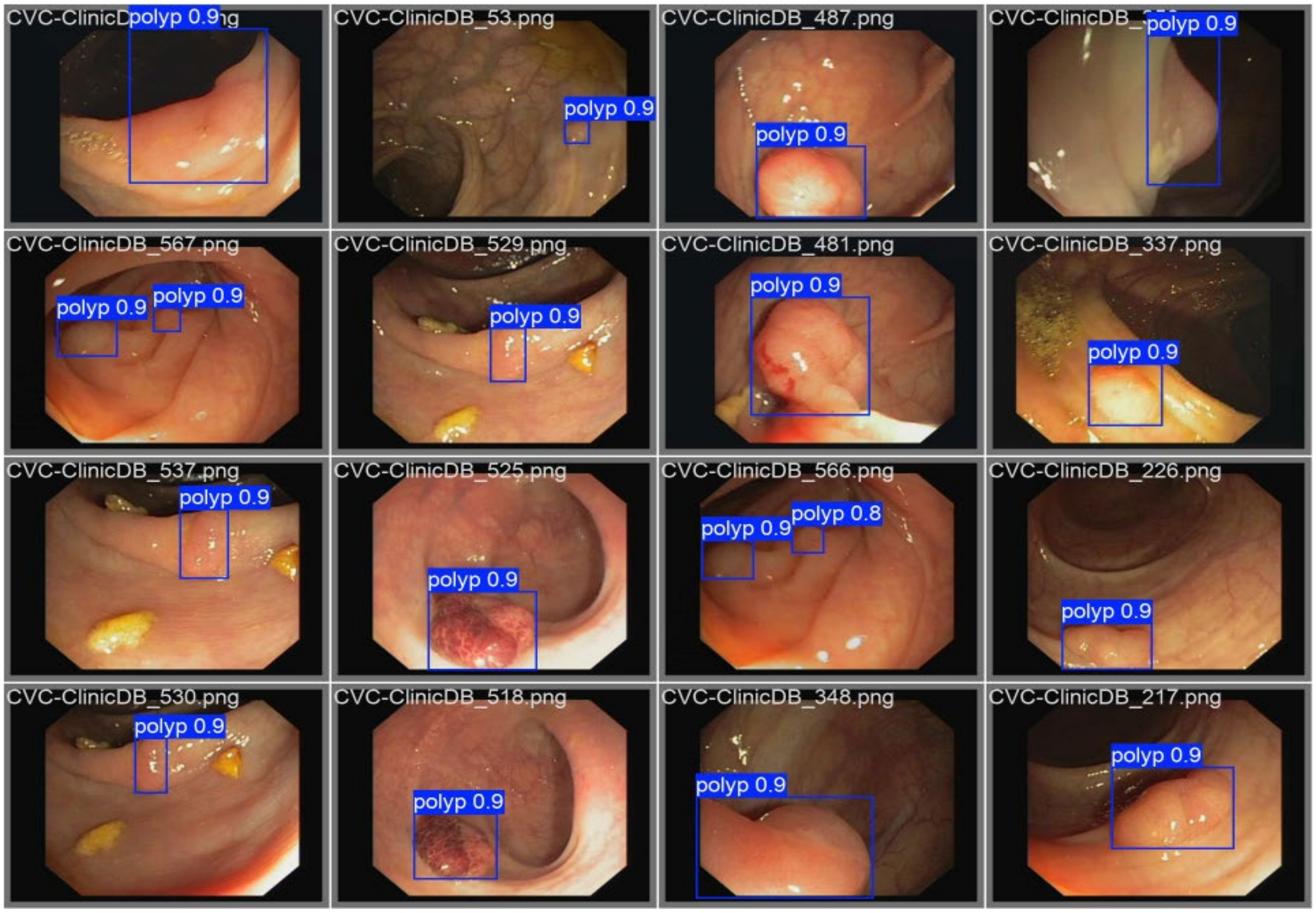

Figure 14. Predicted bounding boxes with confidence scores produced by the deep learning model in batch 1.

Figures 24 and 25 in batch 2 illustrate that this batch presented more visually complex samples, with some polyps partially occluded or embedded within mucosal textures. Despite these challenges, the model maintained high sensitivity, detecting nearly all labeled polyps with minimal deviation in bounding box positioning. Confidence scores remained consistent (~0.8–0.9), indicating the model's confidence in diverse visual contexts. The predictions-maintained alignment with the ground truth, affirming the generalizability of the framework across different patients and image variations.

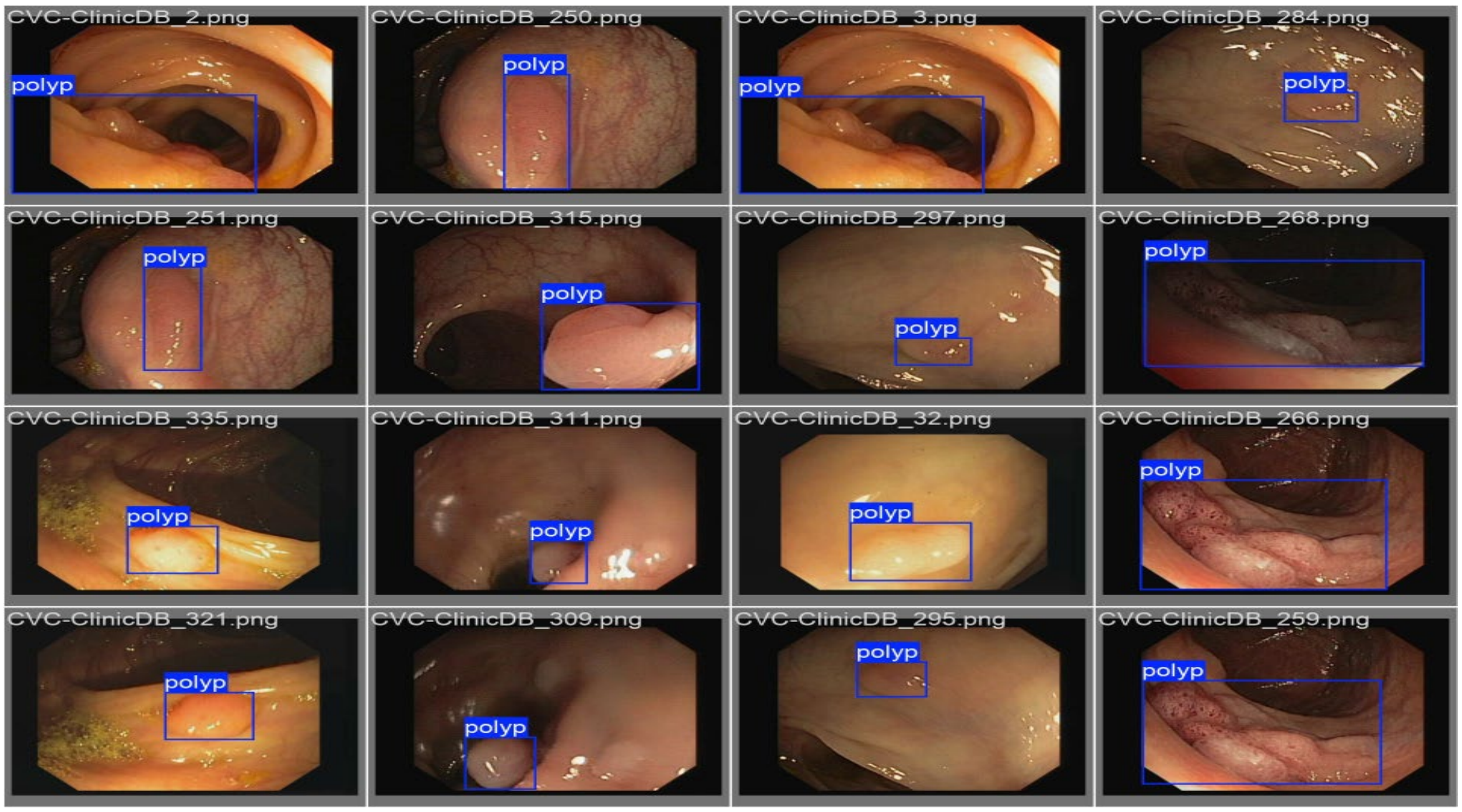

Figure 15. Ground truth bounding boxes indicating the location of polyps in colonoscopy images in batch 2.

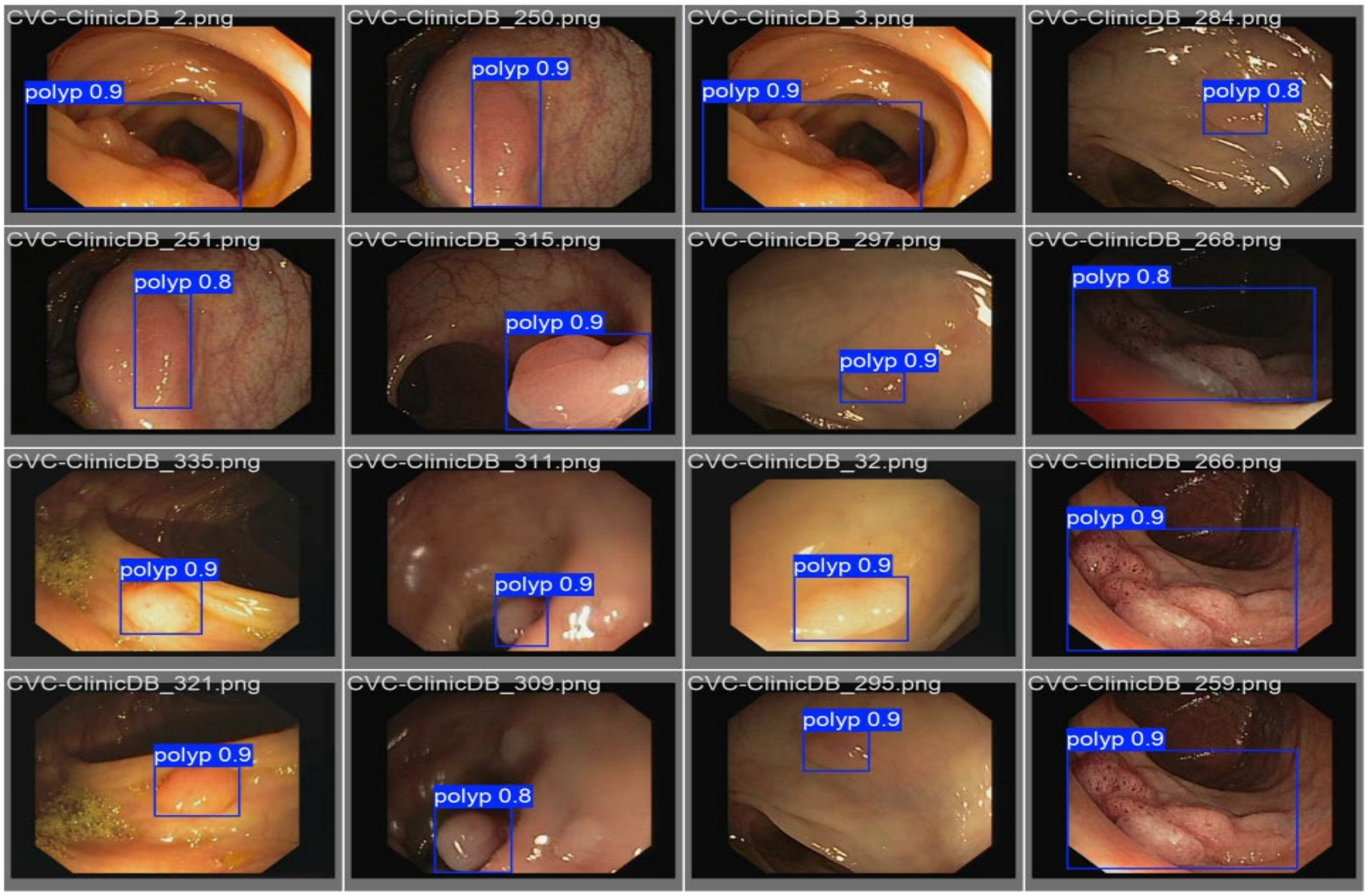

Figure 16. Predicted bounding boxes with confidence scores produced by the deep learning model in batch 2.

In the field of colorectal polyp detection, the LOF-YOLOv11n model emerged as a strong competitor, outperforming several advanced models in terms of recall, precision, and F1-score, as shown in Table 5. This table illustrated the performance of the implemented method in comparison to other approaches, offering valuable insights into its efficacy and potential benefits.
To validate the effectiveness of the proposed LOF + YOLOv11n architecture for polyp detection, it was benchmarked against a comprehensive set of state-of-the-art models using standardized datasets, including Kvasir-SEG, CVC-ClinicDB, CVC-ColonDB, ETIS-LaribPolypDB, and EndoScene. As shown in Table 6, the method achieved the highest overall performance across key evaluation metrics among YOLO-based models. Specifically, it yielded a precision of 95.83%, a recall of 91.85%, and an F1score of 93.48%. In prior works, various techniques were employed to address the inherent complexity of polyp detection, including attention-based modules, lightweight architectures, and patch-based networks. For instance, Wan et al. leveraged YOLOv5 with attention to achieve a respectable 90.7% F1score. Similarly, Butler et al. reported near-saturated performance (94.8%) across all metrics using a lightweight tracker-based architecture.
Traditional CNN-based methods, such as AlexNet [46] and patch-based CNNs [47], lacked full F1 reporting or fell short in balancing precision and recall. Likewise, domain-adaptive models like Faster R-CNN [12] or F-CNNs [48] showed moderate performance (F1 below 87%), indicating limitations in handling complex and varied polyp structures. U-Net variants, including those proposed by Jha et al., demonstrated a precision of only 76.4%, which indicated less effective localization. Additionally, deep convolutional networks tend to operate more slowly than YOLO models, making them unsuitable for faster real-time detection.
While recent YOLO-based architectures [25, 46] improved real-time detection performance, only YOLOv8n approached a comparable F1 score (91.20%) to the proposed method. However, the LOF component in the proposed model acted as a statistical anomaly filter, effectively enhancing background-foreground separation by emphasizing spatial areas with local deviations from the norm namely, potential polyp candidates. Thus, the integration of the Local Outlier Factor (LOF) with YOLOv11n in the proposed method further enhanced detection by improving generalization and minimizing false positives, which proved particularly beneficial in medical imaging, where inter-polyp variability in shape, texture, and size was high. Overall, the proposed model established a new benchmark in the field, with the highest precision, recall, and F1-score among all compared studies, confirming the effectiveness of combining outlier-aware representation with lightweight real-time detection.
Figure 26 presented multiple endoscopic images of the colon in the entire dataset, with green bounding boxes indicating the detected polyps, demonstrating the performance of the polyp detection algorithm under various lighting and anatomical conditions.

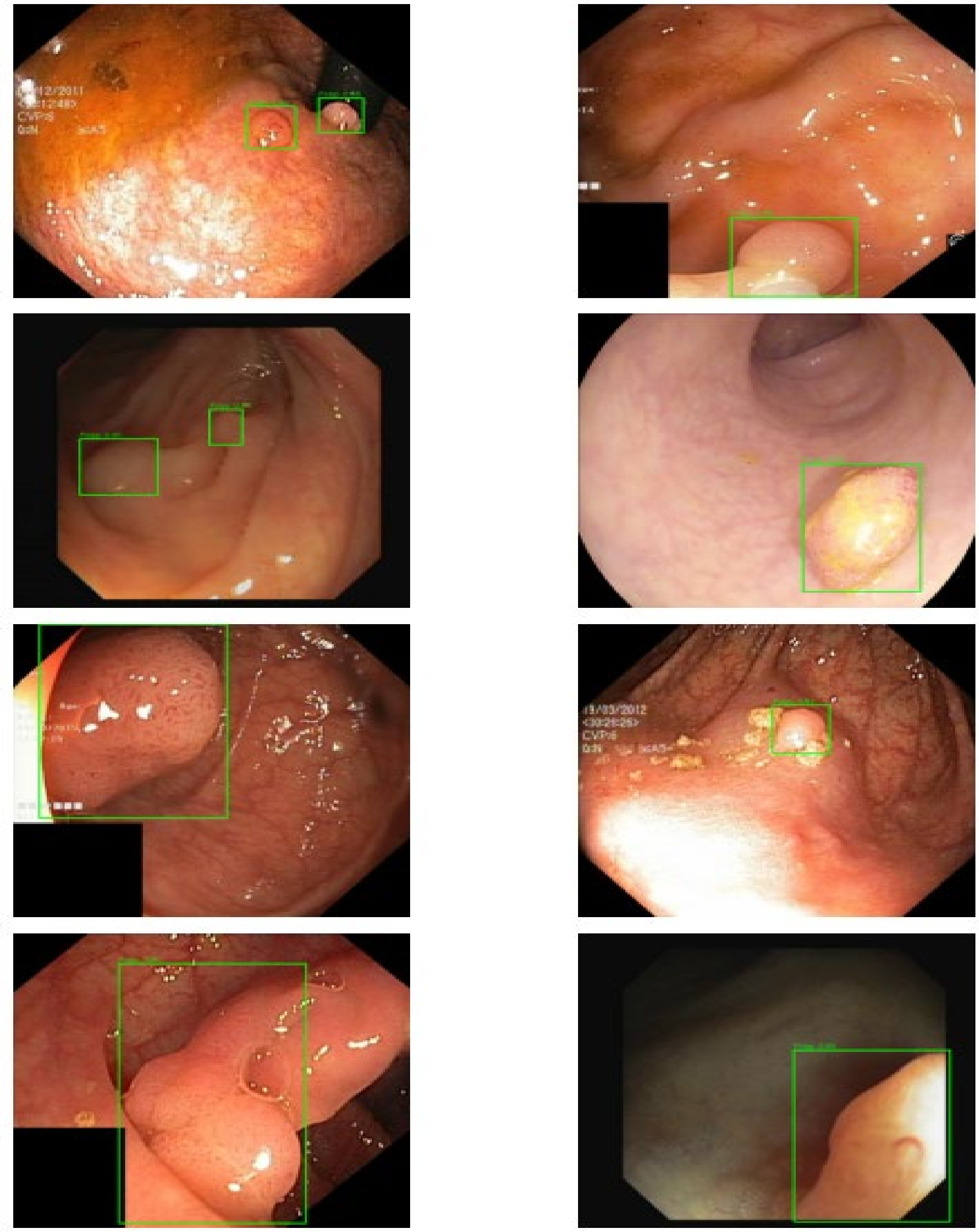

Figure 17. presented the visual results of the proposed method and showed detected polyps with their bounding boxes.

While reducing computing load, single-shot multibox detectors and other designs optimally captured complex features pertinent to polyp identification. Accuracy was maintained during real-time processing due to their efficiency. The LOF-YOLOv11n detector was specifically designed to extract complex visual patterns essential for polyp identification, such as subtle textures and shapes, while maintaining simple computational efficiency. It achieved this goal by combining a high-performance single-stage detection pipeline with a lightweight architecture that minimized additional processing. LOF-YOLOv11n leveraged advanced preprocessing strategies that cleaned training data and enhanced dataset quality to improve YOLO's ability to detect polyps. This adaptability enhanced generalization and robustness, which led to improved performance on unseen data. Therefore, unlike models that required complex preprocessing or extensive data augmentation, LOF-YOLOv11n achieved high recall and precision by default, making it particularly suitable for deployment in heterogeneous real-world clinical environments.
In Figure 27, a subset of single polyp detection findings was shown in the boxes enclosing the identified polyps with green squares in the image.

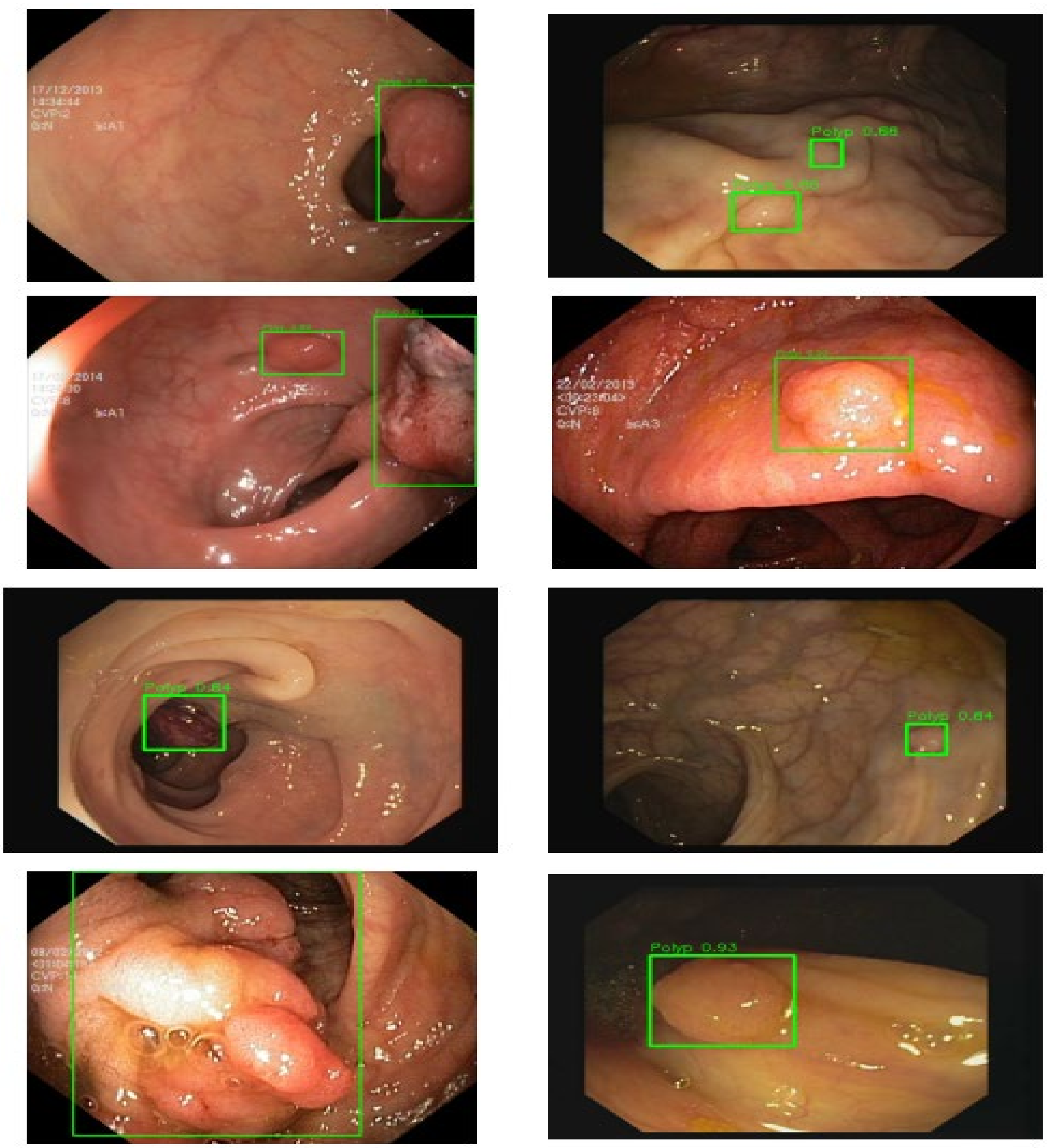

Figure 18. Detected polyps with bounding boxes.

Table 6 presented a comparative analysis of recent deep learning–based methods for colorectal polyp detection alongside our proposed LOF + YOLOv11n model, highlighting its superior balance of precision, recall, and F1-score.

Table 6. Results from relevant and recent studies

| Reference | Model | Dataset | Precision (%) | Recall (%) | F1-score (%) |
|---|---|---|---|---|---|
| Wan et al. [38] | Attentive YOLOv5 | Kvasir-SEG | 91.50 | 89.90 | 90.70 |
| Qadir et al. [47] | MDeNetplus (F-CNN) | CVC-ColonDB | 88.35 | 91.00 | 89.65 |
| Shin et al. [48] | CNN (Patch-based) | CVC-ClinicDB, ETIS-LARIB, ASU-Mayo | 92.71 | 90.82 | – |
| Zhou et al. [49] | Geometric & Texture (VCE) | Video Capsule Endoscopy | – | – | – |
| Sahoo et al. [50] | YOLOv11n | Kvasir-SEG | 91.76 | 90.29 | 91.01 |
| Butler et al. [51] | Lightweight Tracker + Detection | CVC-ClinicDB | 94.80 | 94.80 | 94.80 |
| Jha et al. [52] | ResUNet++ + CRF + TTA | Kvasir-SEG / CVC-ClinicDB | 93.10 / 88.20 | 91.40 / 86.50 | 92.30 / 87.10 |
| Yeung et al. [53] | Lightweight U-Net + Focus | Kvasir-SEG / CVC-ClinicDB | 91.00 / 94.10 | – | – |
| Lalinia et al. [25] | YOLOv8m / YOLOv8n | Kvasir-SEG, CVC-ClinicDB, CVC-ColonDB, ETIS, EndoScene | 95.60 / 95.20 | 91.70 / 90.30 | 92.40 / 91.20 |
| Our Proposed Method | Kvasir-SEG, CVC-ClinicDB, CVC-ColonDB, ETIS, EndoScene | LOF + YOLOv11n | 95.83% | 91.85% | 93.48% |

As shown in Table 7, the method achieved the highest Precision (95.83%), Recall (91.85%), and F1 Score (93.48%), indicating a robust balance between minimizing false positives and effectively identifying true polyp instances. This made it particularly suitable for clinical settings where both sensitivity (recall) and specificity (precision) were critical. Moreover, the method achieved a mean Average Precision at IoU 0.5 (mAP@0.5) of 96.48%, slightly outperforming closely related models such as YOLOv11n (96.25%) and YOLOv10 (95.16%). This suggested a marginal but valuable improvement in detecting polyps at standard intersection-over-union (IoU) thresholds.

**Table 7.** Average Results and Standard Deviation of the 5-Fold Cross-Validation Method

| Method | Precision | Recall | F1-Score | mAP@0.5 | mAP@0.5:0.95 |
|---|---|---|---|---|---|
| **LOF-yolov11n proposed Method** | $95.83 \pm 1.0$ | $91.85 \pm 1.10$ | $93.48 \pm 0.88$ | $96.48 \pm 1.01$ | $77.75 \pm 1.05$ |

### 5.1. Limitations

Although the YOLO11M-LOF framework provided a new and effective integration of object detection with anomaly detection using the Local Outlier Factor (LOF), several limitations must also be considered. One of the main limitations was the availability and diversity of high-quality, annotated datasets. Many publicly available datasets contained a limited number of anomaly samples or lacked sufficient diversity among patients and imaging conditions. This limits the model's ability to generalize to broader clinical or real-world scenarios.

Additionally, the performance of LOF, which is sensitive to data distribution, may vary in areas where the feature space is less structured or contains noise. While this model achieved strong results in controlled environments, its reliability in more complex or unstructured settings requires further investigation.

Moreover, it may encounter hardware limitations when scaled to larger datasets. This was primarily due to increased computational demands associated with processing high-resolution images and larger volumes of data. As the dataset size grew, requirements for memory, storage, and processing power also increased, potentially exceeding the capacity of standard hardware configurations.

## 6. Conclusion and future work

This research introduced a streamlined and practical method for detecting colorectal polyps in colonoscopy images, focusing on computational efficiency and diagnostic accuracy. By incorporating a local outlier factor for anomaly filtering before training, the proposed system was able to exclude misleading data samples, which in turn led to more stable learning and improved overall detection outcomes. Utilizing the YOLO-v11n model, selected for its balance between speed and precision, the framework was validated across five distinct datasets, with experimental results confirming its strong performance in various metrics such as precision, recall, and mAP.

Despite these encouraging results, the method presents opportunities for further development. Future work may investigate dynamic outlier detection techniques that adapt to varying data distributions or clinical settings. Additionally, integration of temporal cues in video-based colonoscopy or deployment on embedded medical hardware could enhance its utility in practice. Expanding the system's capability to differentiate between polyp types or detect early-stage lesions could also increase its diagnostic value. Overall, the findings highlight the benefits of combining data preprocessing with efficient model design for reliable and real-time medical image analysis.

### Author statements

**Ethical approval**
None required as this is a secondary data analysis.

**Funding**
This research received no external funding.

**Competing interest**
There are no conflicts of interest to declare.

**Data availability**
The datasets are available at:
http://vi.cvc.uab.es/colon-qa/cvccolondb/
https://polyp.grand-challenge.org/CVCClinicDB/
https://datasets.simula.no/kvasir-seg/
https://polyp.grand-challenge.org/ETISLarib/

http://adas.cvc.uab.es/endoscene